\documentclass[10pt,letterpaper]{article}
\usepackage{arxiv}
\usepackage{preprint}
\usepackage{graphicx,amsmath,amssymb}
\usepackage{algorithm,algpseudocode}
\usepackage{booktabs,array,tabularx,colortbl}
\usepackage{xcolor}
\definecolor{PhysScopeBlueTint}{HTML}{EBF2F8}
\definecolor{PhysScopeLink}{HTML}{245878}
\usepackage[font=small,labelfont=bf,labelsep=period]{caption}
\usepackage[numbers,sort&compress]{natbib}
\usepackage{chapterbib}
\usepackage{microtype}
\ifdefined\PhysScopePromptStyleLoaded\expandafter \fi
\newcommand{\PhysScopePromptStyleLoaded}{}

\usepackage{listings}
\usepackage{xcolor}
\usepackage{needspace}

\definecolor{physScopePromptBlue}{RGB}{65,106,118}
\lstdefinestyle{physScopePrompt}{
  basicstyle=\ttfamily\small,
  columns=fullflexible,
  keepspaces=true,
  breaklines=true,
  breakatwhitespace=false,
  breakindent=0pt,
  showstringspaces=false,
  showspaces=false,
  showtabs=false,
  numbers=none,
  frame=none,
  aboveskip=4pt,
  belowskip=4pt,
  morecomment=[l]{\#},
  commentstyle=\color{physScopePromptBlue},
  mathescape=false,
  texcl=false
}
\lstdefinestyle{physScopeSchema}{
  style=physScopePrompt,
  frame=tb,
  framerule=0.3pt,
  aboveskip=5pt,
  belowskip=5pt
}
\newcounter{physscopeprompt}
\newcommand{\PhysPromptHeading}[2]{%
  \par\addvspace{7pt}%
  \Needspace{5\baselineskip}%
  \refstepcounter{physscopeprompt}\label{#1}%
  \vbox{%
    \hrule height 0.6pt\relax
    \vskip3pt
    \raggedright\noindent\textbf{Prompt~\thephysscopeprompt.}~#2\par
    \vskip3pt\hrule height 0.3pt\relax
  }%
  \nobreak
}
\newcommand{\PhysPromptEnding}{%
  \par\nobreak\hrule height 0.3pt\relax
  \addvspace{7pt}%
}
\lstnewenvironment{PhysPrompt}[2]
  {\ifdefined\nolinenumbers\nolinenumbers\fi
   \PhysPromptHeading{#1}{#2}\lstset{style=physScopePrompt}}
  {\PhysPromptEnding}

\newenvironment{PhysScopeWideAppendix}{\par\raggedbottom}{\par}

\usepackage[breaklinks,colorlinks,allcolors=PhysScopeLink]{hyperref}
\hypersetup{
  pdftitle={PhysScope: Agentic Multimodal Physics Reasoning with Grounded Visual Memory},
  pdfauthor={Haoyang Liu, Kecheng Liang, Xinyu Li, Weixing Chen, Yang Liu}
}
\renewcommand{\shorttitle}{PhysScope}
\title{PhysScope: Agentic Multimodal Physics\\Reasoning with Grounded Visual Memory}
\author{%
  \textbf{Haoyang Liu}\textsuperscript{1}\quad
  \textbf{Kecheng Liang}\textsuperscript{2}\quad
  \textbf{Xinyu Li}\textsuperscript{2}\quad
  \textbf{Weixing Chen}\textsuperscript{2}\quad
  \textbf{Yang Liu}\textsuperscript{2,$\dagger$}\\[5pt]
  {\small\textsuperscript{1}Dalian University of Technology}\\[2pt]
  {\small\textsuperscript{2}Sun Yat-sen University}\\[4pt]
  {\small\textsuperscript{$\dagger$}Corresponding author: \href{mailto:liuy856@mail.sysu.edu.cn}{\texttt{liuy856@mail.sysu.edu.cn}}}\\[3pt]
  {\small Code will be released soon.}%
}
\date{}
\begin{document}
\maketitle

\begin{abstract}
Vision-language models (VLMs) can correctly interpret physics diagrams yet apply constraints outside the states or conditions in which they hold, causing premise errors to propagate through subsequent calculations.
We introduce \textbf{PhysScope}, an inference-time visual harness that supports self-correction of these scope mismatches without updating model weights.
PhysScope enables the model to link physical scopes directly to its visual observations and explicitly constructs derived views to support reasoning.
The harness imposes explicit constraints on state distinctions, conditions, and physical scopes during model reasoning, and checks agent-proposed revisions against sources and scopes.
Accepted revisions invalidate registered dependents, thereby maintaining scope-aware working memory.
On PhysAlign, PhysScope achieves higher solving scores than direct solving across all six evaluated VLMs. With Gemini-3.8-flash, it outperforms Canvas-CoT by 9.34 percentage points and substantially improves solving accuracy on the out-of-distribution (OOD) PhysUniBench benchmark.
Ablation studies support explicit scope management and dependency maintenance as design principles for multimodal physics reasoning agents.
\end{abstract}
\section{Introduction}
\label{sec:introduction}

Multimodal physics reasoning requires vision-language models (VLMs) to perform multistep problem solving from text, diagrams, and physical conditions.
A key challenge is maintaining the applicability of the premises used in reasoning.
The same object may be subject to different constraints at different stages, and a relation valid in one state may not apply to a subsequent process.
Even when a model correctly recognizes the objects, numerical values, and scene, using a constraint outside the states or conditions in which it holds can propagate errors through subsequent calculations.
Agents for physics reasoning must therefore maintain premise scopes throughout problem solving.

\begin{figure}[htbp]
    \centering
    \includegraphics[width=1\textwidth]{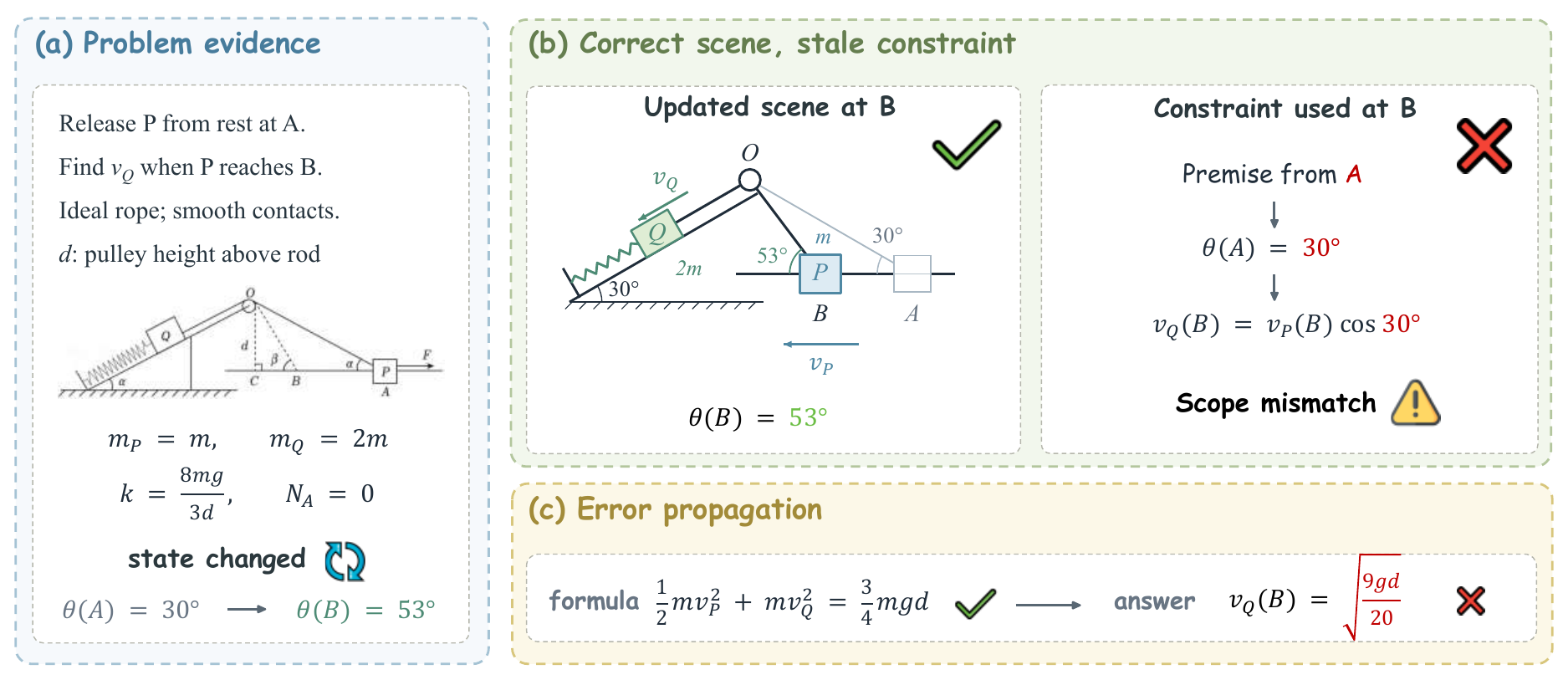}
    \caption{\textbf{The motivation: correct scene, stale constraint.}
    \textbf{(a)} A physics problem from PhysAlign~\cite{liang2026physalignbenchmarkevidencegroundedrole}.
    \textbf{(b)} The updated scene depicts $\theta(B)=53^\circ$, but the calculation at $B$ reuses the $30^\circ$ angle from state $A$.
    \textbf{(c)} This scope mismatch propagates through a correct energy equation to an incorrect final speed.
    This failure illustrates why state updates must propagate to premise use and dependent calculations.}
    \label{fig:physscope_motivation}
\end{figure}

Figure~\ref{fig:physscope_motivation} illustrates how a mismatch in premise applicability can lead to an incorrect solution.
The rope segment makes an angle of $30^\circ$ in state $A$ and $53^\circ$ in state $B$.
Although the model correctly represents the scene at $B$, it substitutes the $30^\circ$ angle from $A$ into the velocity constraint at $B$, yielding an incorrect speed.
We define the \emph{scope} of a premise by the entities it concerns, the states in which it applies, and the conditions under which it holds.
In this example, the angle at $A$ remains valid historical information, but it cannot be used as the current angle at $B$. The velocity relation must also satisfy the corresponding rope-length constraint.
This \emph{scope mismatch} shows that an accurate visual representation does not automatically ensure correct use of premises: the link between each premise and its applicability conditions must be preserved in the derivation itself.

External representations and visual memory provide a basis for supporting such reasoning.
Visual Sketchpad~\cite{hu2024visual} supports visual reasoning through intermediate sketches, Canvas-CoT~\cite{sun2026canvas} supports state revision through an editable canvas and visual checks, and VISTA~\cite{vista2026} preserves original visual observations for later inspection during reasoning.
These capabilities allow models to use external information to assess and revise their understanding of a problem.
For physics problems, these adjustments must also be reflected in premise use: whether a relation applies to the current state, and which conclusions remain supported after it is revised, directly affect subsequent calculations.
An agent harness for physics reasoning must therefore connect the maintenance of visual information with updates to the premises used for solving.

To this end, we introduce \textbf{PhysScope}, an inference-time visual harness that requires no updates to model weights.
PhysScope organizes working memory around physical premises with explicit scopes, generates textual feedback and derived views from the same premises, and supports revisions to reasoning grounded in source evidence.
The framework connects visual observations, premise use, and updates to derivations, supporting the model in identifying scope mismatches and revising the relevant derivations during problem solving.

PhysScope's innovation lies in explicitly managing the applicability of physical premises and the consequences of their revision through the harness.
Canvas-CoT~\citep{sun2026canvas} organizes reasoning around editable structured states, and VISTA~\citep{vista2026} preserves original observations for visual reinspection. PhysScope maintains the premises used for solving by tracking their scopes and derivation dependencies.
Under this design, whether a premise applies to the current state becomes checkable, and accepted premise corrections invalidate registered derivations that depend on the superseded content.
This preserves physical relations that remain valid in their respective states and identifies which conclusions need to be re-established.
Updates to visual memory can therefore influence subsequent calculations and provide targeted feedback for model self-correction.

We assess this design through evaluations across models and experiments examining its mechanisms.
On PhysAlign~\cite{liang2026physalignbenchmarkevidencegroundedrole}, PhysScope achieves higher solving scores than direct solving across all six evaluated VLMs.
With Gemini-3.8-flash~\citep{googledeepmind2026gemini38flash}, it exceeds Canvas-CoT~\citep{sun2026canvas} by 9.34 percentage points and improves problem-solving accuracy in external evaluation on PhysUniBench~\cite{wang2025physunibench}.
Controlled comparisons with shared physical content also show gains in solving scores from adding derived views.
Ablation studies show that removing scope management leaves content recognition accuracy unchanged but markedly reduces solving scores, removing dependency maintenance also lowers solving scores.
These results support explicitly maintaining premise applicability and the consequences of revision in multimodal physics reasoning.

Our main contributions are as follows:
\begin{itemize}
    \item \textbf{Scope-aware physical working memory.}
    We organize working memory around physical premises associated with entities, states, and applicability conditions.
    This makes premise applicability an explicit responsibility of the harness and supports checking the scope within which visual information is used in subsequent derivations.

    \item \textbf{Visual feedback consistent with physical premises.}
    We generate textual feedback and derived views from shared premises, preserving applicability conditions and unknowns while presenting differences between states.
    This connects visual observations to applicable premises and helps the model identify scope mismatches.

    \item \textbf{Controlled, dependency-aware memory revision.}
    We combine source and scope checks and, upon accepting a premise correction, invalidate registered derivations that depend on the previous premise.
    This identifies conclusions that need to be re-established and allows local corrections to propagate to subsequent problem solving.
\end{itemize}
\section{Related Work}
\label{sec:related_work}

\paragraph{Multimodal physics reasoning and role grounding.}
Multimodal benchmarks evaluate how models combine visual evidence with domain knowledge.
MathVista~\cite{lu2024mathvista} studies mathematical reasoning in visual contexts.
In physics, SeePhys~\cite{xiang2026seephys} spans educational levels and diagram types, PhysUniBench~\cite{wang2025physunibench} targets undergraduate problems, and PhyX~\cite{shen2025phyx} covers diverse physical domains and reasoning types.
PhysReason~\cite{zhang2025physreason} evaluates final answers and intermediate steps.
Targeted diagnostics separate perception from physical interpretation: PhysAlign~\cite{liang2026physalignbenchmarkevidencegroundedrole} distinguishes content recognition from role grounding via localized probes and controlled variants, while SeePhys Pro~\cite{xiang2026seephyspro} transfers critical information from text to images across semantically equivalent variants.
For solving, Zhang et al.~\cite{zhang2026decoupled} train models to separate physical modeling from mathematical execution via supervised fine-tuning and rubric-based reinforcement learning.
Physics Supernova~\cite{qiu2025physicssupernovaaiagent} integrates image analysis and answer review into an agent workflow with access to computational tools.
PhysScope focuses on premise use during inference, linking explicit object, state, and condition scopes to dependent calculations and subsequent revisions.

\noindent\textbf{Visual reasoning with external representations.}
External representations make intermediate reasoning accessible through programs, image regions, and editable diagrams.
VISPROG~\cite{gupta2023visual} and ViperGPT~\cite{suris2023vipergpt} compose visual and language modules into executable programs.
Visual CoT~\cite{shao2024visual} focuses successive reasoning turns on relevant image regions, while Visual Sketchpad~\cite{hu2024visual} lets models draw and inspect intermediate visual artifacts.
Canvas-CoT~\cite{sun2026canvas} maintains a mutable document object model (DOM) and uses a critic to compare its rendered state with the original image.
PhysScope organizes intermediate physical content as claims with sources, explicit scopes, and versioned dependencies.
Derived views preserve registered conditions and unknowns, and both text and visual feedback come from Scope-Aware Working Memory.
This supports comparisons that examine visual organization while matching the physical content available to the solver.

\noindent\textbf{Agent memory and evidence-grounded revision.}
Agent memory enables revisiting evidence and prior reasoning.
VISTA~\cite{vista2026} retains original visual observations for active retrieval and input reorganization.
ReMAP~\cite{jiang2026remaprestoringperceptualcycle} combines global and local latent visual memories with a learned policy governing when and which memory to access.
For video plausibility assessment, PhysMRV~\cite{wang2026physmrv} retrieves scene descriptions, physical-event graphs, and rule summaries from training videos to support a frozen VLM.

For revision, Self-Refine~\cite{madaan2023self} uses the same model to critique and revise its outputs, whereas Reflexion~\cite{shinn2023reflexion} stores reflections on task feedback for subsequent trials.
EVPV~\cite{wang2026groundingscoreexplicitvisual} matches explicit visual premises against independently extracted constraints to calibrate process rewards and candidate ranking.
Classical truth-maintenance systems~\cite{doyle1979truth} record justifications and revise beliefs as their assumptions change.
PhysScope maintains Scope-Aware Working Memory within the current problem.
Local revisions are checked against recorded sources and scopes, accepted changes invalidate registered dependent facts, equations, and views.
Semantic repair uses an updated snapshot when an unresolved issue is answer-relevant, actionable, and within budget, protocol-triggered regeneration is accounted for separately.

\section{Method}
\label{sec:method}

\begin{figure}[htbp]
    \centering
    \includegraphics[width=1\textwidth]{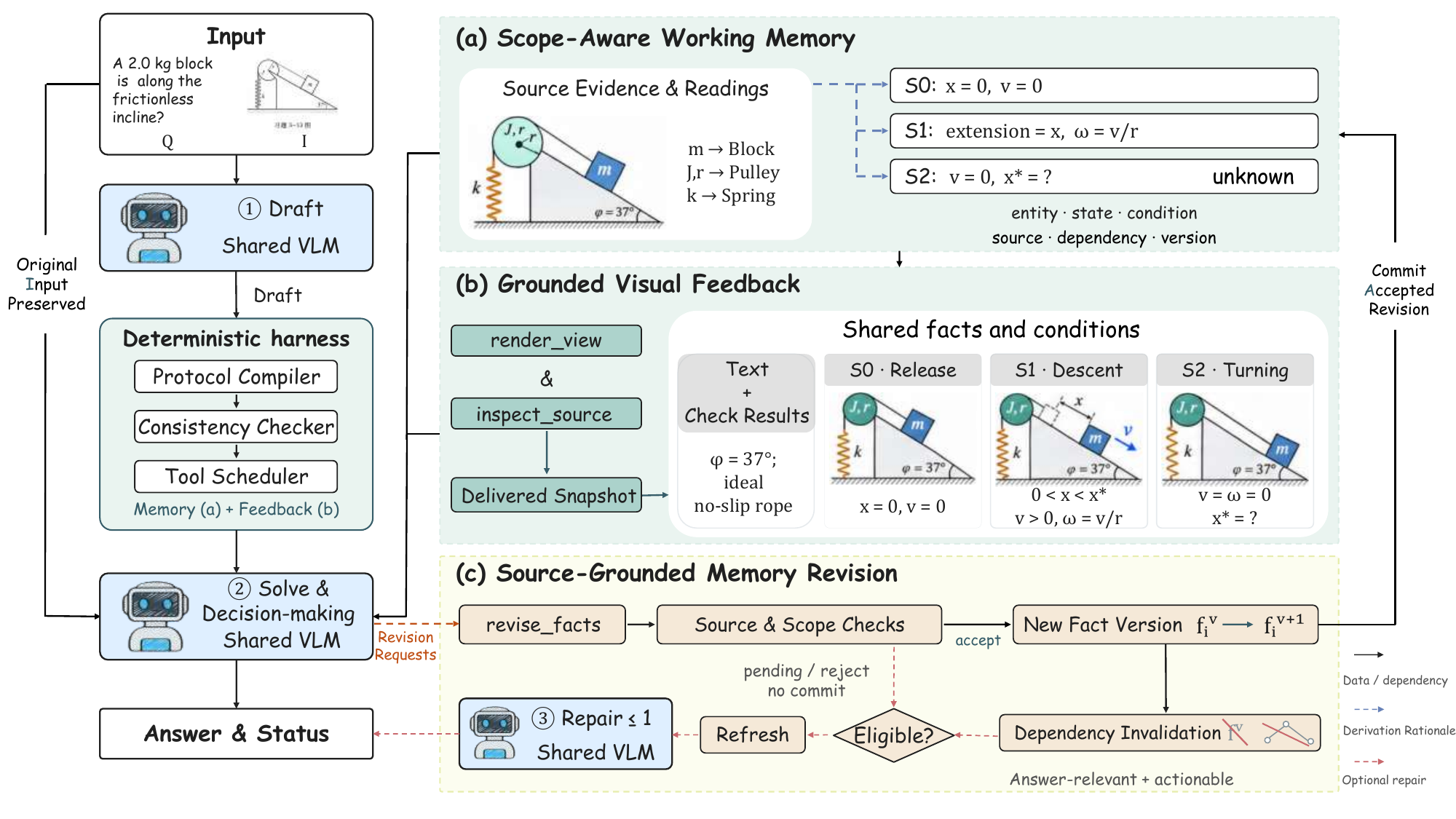}
    \caption{\textbf{Overview of PhysScope.}
    The shared VLM performs Call~1 (\emph{Draft}) and Call~2 (\emph{Solve \& Decision-making}).
    The deterministic harness manages three connected components:
    \textbf{(a)} Scope-Aware Working Memory associates premises with entities, states, conditions, sources, and versioned dependencies.
    \textbf{(b)} Grounded Visual Feedback presents the same registered physical content in text and graphics.
    A \emph{Delivered Snapshot} records the memory, source evidence, check feedback, and views supplied to a model call.
    \textbf{(c)} Source-Grounded Memory Revision checks proposed changes, commits accepted replacements, and invalidates registered dependents.
    Pending or rejected revisions leave active versions unchanged.
    Dashed arrows mark the optional \emph{Repair} path (Call~3). Semantic repair and protocol recovery share one additional call.}
    \label{fig:physscope_overview}
\end{figure}

PhysScope maintains physical premises as source-linked records with explicit scopes and versioned dependencies.
Its design connects premise applicability, visual feedback, and memory revision: the same registered content supports text and views, while accepted premise replacements invalidate registered dependents.
A shared VLM interprets the problem and derives physical content, a deterministic harness checks declared references and controls memory updates.
Figure~\ref{fig:physscope_overview} summarizes this interaction.

\subsection{Agent--harness inference protocol}
\label{sec:inference_protocol}

Given a statement $Q$ and diagram $I$, the core protocol runs Call~1 (\emph{Draft}) and Call~2 (\emph{Solve \& Decision-making}) on the same VLM $\mathcal{L}$.
\begin{equation}
\label{eq:physscope_protocol}
\begin{aligned}
\widetilde{\mathcal{M}} &= \mathcal{L}(Q,I,\mathrm{draft}), \\
(\mathcal{M},\mathcal{O}_2) &=
\mathcal{H}_{\mathrm{prepare}}(Q,I,\widetilde{\mathcal{M}}), \\
(\hat{y}_2,\Delta_2) &=
\mathcal{L}(Q,I,\mathcal{O}_2,\mathrm{solve}).
\end{aligned}
\end{equation}
Call~1 drafts $\widetilde{\mathcal{M}}$ by recording entities, source readings, relations, candidate states, and unknowns.
The Protocol Compiler assigns identifiers and versions while preserving declared scopes and dependencies.
The Consistency Checker evaluates source delivery, version validity, and scope compatibility. The Tool Scheduler selects source inspections and supported views.
These components prepare memory $\mathcal{M}$ and the Delivered Snapshot $\mathcal{O}_2$.
Given $Q$, $I$, and $\mathcal{O}_2$, Call~2 returns a provisional answer $\hat{y}_2$ and an incremental submission $\Delta_2$ containing new conclusions, revision requests, and unresolved issues.
Key equations and answer claims specify applicable states and direct dependencies.

The agent supplies physical content, source locations, and dependencies.
The harness executes tools and determines which proposed updates to commit.
Original inputs remain available throughout inference.
The harness may authorize one additional call (Call~3) for an eligible semantic issue or recoverable protocol failure (Algorithm~\ref{alg:physscope}).
Reference answers, ground-truth entity assignments, and analysis-only labels are excluded from inference and scheduling.

\subsection{Scope-Aware Working Memory}
\label{sec:physical_memory}

A premise may remain valid for an earlier state while being inapplicable to the current derivation.
PhysScope therefore records the scope of each premise separately from its content.
Memory distinguishes immutable source evidence, model reading claims, working facts, and unresolved issues.
Each fact version is represented as
\begin{equation}
\label{eq:physical_fact}
f_i^{(v)} =
\bigl(p_i,e_i,s_i,c_i,\rho_i,D_i,\tau_i\bigr),
\end{equation}
where $p_i$ is the content, $e_i$, $s_i$, and $c_i$ specify the associated entities, state scope, and applicability conditions, $\rho_i$ records sources, $D_i$ lists direct dependencies with exact versions, and $\tau_i$ denotes claim status.
Scopes distinguish temporal phases, physical regimes, and spatial regions.
Claim statuses are \texttt{observed}, \texttt{derived}, \texttt{hypothesis}, and \texttt{unknown}.
An observed claim is a model reading of a source, a derived claim must identify its supporting premises.

Figure~\ref{fig:physscope_overview}(a) illustrates these distinctions.
The parameters $m$, $(J,r)$, and $k$ belong to the block, pulley, and spring, respectively.
The readings $x=0$ and $v=0$ apply to release state $S_0$.
In descent state $S_1$, $\omega=v/r$ holds under ideal-rope and no-slip conditions, while the turning position $x^{\ast}$ in $S_2$ remains unknown.
State evolution adds scoped records without overwriting valid historical readings.
An omitted scope does not imply global applicability.
Appendix~\ref{sec:appendix_memory_schema} details applicability checks, typed temporal contracts, and the use of historical initial conditions.

The Consistency Checker evaluates declared references, source delivery, exact versions, and scope compatibility.
Each check records its target, rule, outcome, and reason, with outcomes \texttt{pass}, \texttt{fail}, \texttt{unknown}, or \texttt{unsupported}.
Optional physical checks cover predefined relations with explicit premises.
A passing result establishes consistency with the executed rules.
This representation makes declared premise applicability available to both programmatic checks and subsequent visual feedback.

\subsection{Grounded Visual Feedback}
\label{sec:visual_feedback}

To expose scope distinctions during reasoning, PhysScope generates textual feedback and derived views from the same registered physical content.
The operation \texttt{inspect\_source} retrieves a specified text span or original image region.
The operation \texttt{render\_view} constructs a view from selected facts and states:
\begin{equation}
\label{eq:physical_rendering}
\mathcal{V} =
R\bigl(\mathcal{F}_{\mathrm{sel}},
       \mathcal{S}_{\mathrm{sel}},\sigma\bigr),
\end{equation}
where $\sigma$ specifies the view type and layout.
Every displayed physical relation, condition, and check result has a corresponding textual record.
Rendering introduces no additional physical inference: rule-derived content must first be registered with its supporting premises and rule.

\begin{figure}[htbp]
    \centering
    \includegraphics[width=1\textwidth]{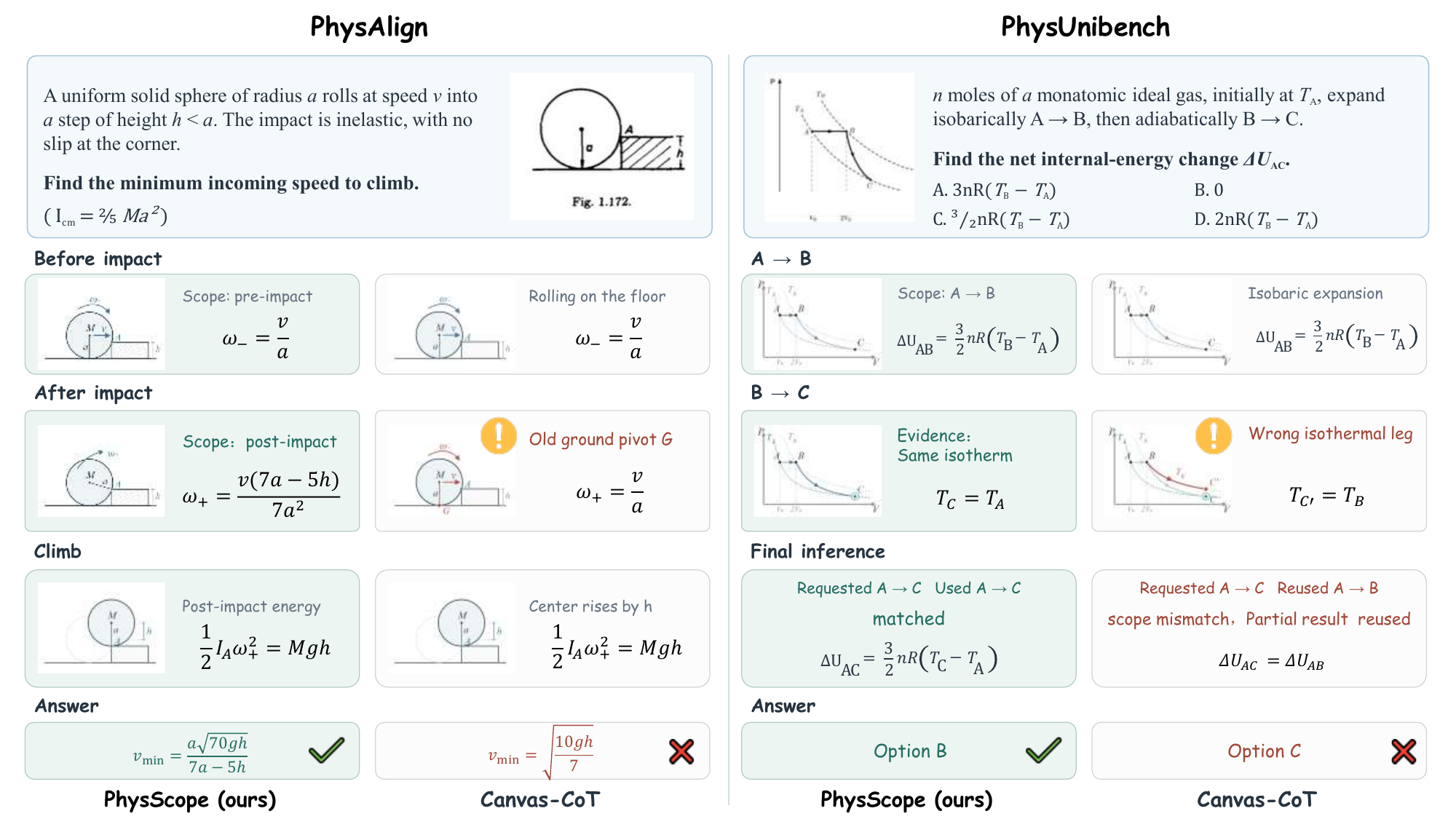}
    \caption{\textbf{Scope distinctions in physical reasoning.}
    Examples from PhysAlign~\cite{liang2026physalignbenchmarkevidencegroundedrole} (left) and PhysUniBench~\cite{wang2025physunibench} (right).
    Each example compares recorded PhysScope and Canvas-CoT~\citep{sun2026canvas} trajectories, intermediate diagrams are explanatory redraws.
    In the impact example, the Canvas-CoT trajectory carries the pre-impact rolling relation across the collision, while the PhysScope trajectory separates impact dynamics from climbing energy.
    In the gas example, the Canvas-CoT~\citep{sun2026canvas} trajectory treats the adiabatic leg as isothermal and reuses $\Delta U_{AB}$ for the full process, the PhysScope trajectory uses $T_C=T_A$ to obtain $\Delta U_{AC}=0$.
    Red boxes mark scope mismatches, green boxes highlight the relevant state or endpoint evidence.}
    
    \label{fig:physscope_trajectories}
\end{figure}

Views preserve the conditions and unresolved quantities of their source records.
For example, $v=\omega=0$ at $S_2$ in Fig.~\ref{fig:physscope_overview}(b) denotes an instantaneous turning point, it does not establish static equilibrium, and $x^{\ast}$ remains unknown.
Conditional branches retain their status in text and graphics, and missing defining conditions prevent a definite depiction.
Single-state tasks use local relation views.
Entity identities and coordinate conventions remain consistent, while unsupported magnitudes are represented schematically.

The package actually supplied to model call $t$ is its \emph{Delivered Snapshot}:
\begin{equation}
\label{eq:delivered_snapshot}
\mathcal{O}_t =
\bigl(\mathcal{M}_t^{\mathrm{in}},
      \mathcal{E}_t,\mathcal{C}_t,\mathcal{V}_t\bigr).
\end{equation}
Here, $\mathcal{M}_t^{\mathrm{in}}$ is the exposed versioned memory, $\mathcal{E}_t$ the delivered source evidence, $\mathcal{C}_t$ the check feedback, and $\mathcal{V}_t$ the delivered views.
Each view records its dependent fact versions and conditions, and delivery records identify the receiving call.
A generated image or file path alone does not establish that the model received visual input.
These records allow revision checks to distinguish available evidence from artifacts that were only generated.

Figure~\ref{fig:physscope_trajectories} makes two scope distinctions concrete.
In the impact problem, the rolling relation $\omega^{-}=v/a$ applies before impact.
Post-impact motion uses corner $A$ as the pivot, and the subsequent climbing calculation must use the post-impact angular velocity.
In the gas problem, the adiabatic condition governs $B\rightarrow C$, whereas $T_C=T_A$ relates the endpoints of $A\rightarrow C$ and yields $\Delta U_{AC}=0$.
The first example separates successive physical stages, the second distinguishes a local process from an endpoint relation.
Both illustrate why premise applicability must be preserved from intermediate representations through the final equations.

\subsection{Source-Grounded Memory Revision}
\label{sec:revision}

Correcting a premise requires reconsidering the conclusions supported by its previous version.
PhysScope distinguishes new derivations and state-specific additions from replacements of existing premises.
Additions extend memory, replacements use \texttt{revise\_facts} and undergo source, reference, version, and scope checks (Fig.~\ref{fig:physscope_overview}(c)).
The harness accepts, rejects, or defers each request.
Revising a source reading requires matching source reinspection evidence delivered to the proposing call.
Without this evidence, the request remains \texttt{pending}, and its candidate content cannot become an established premise in a rendered view.
Pending and rejected requests leave active fact versions unchanged.

An accepted replacement commits a new version $f_i^{(v+1)}$ while preserving source evidence and revision history.
Registered dependencies form a directed graph $\mathcal{D}$ from premises to dependent records, including facts, equations, and views.
Replacing $f_i^{(v)}$ invalidates
\begin{equation}
\label{eq:dependency_invalidation}
\mathcal{I}_i^{(v)} =
\{f_i^{(v)}\}
\cup
\operatorname{Desc}_{\mathcal{D}}\!\bigl(f_i^{(v)}\bigr).
\end{equation}
Affected derivations must be re-established from current premises.
Stale references are not automatically redirected to the new version.
Propagation follows registered dependencies but cannot detect omitted ones.

For example, correcting a mass reading invalidates a registered energy equation and any views that depend on that version, even when the equation's symbolic form is unchanged.
Unrelated records remain usable.
Invalidation identifies which derivations have lost valid support without deriving their replacements.

A committed change can affect subsequent reasoning only if updated memory and feedback reach another model call.
After Call~2, the harness processes revision requests and delivers accepted changes in $\mathcal{O}_3$ if Call~3 occurs.
This distinguishes committed updates from their delivery to subsequent inference.
Appendix~\ref{sec:appendix_activation} defines the trajectory-level activation criterion.

\subsection{Tool scheduling and bounded repair}
\label{sec:scheduling_repair}

The Tool Scheduler selects source inspections and supported views under a policy fixed before inference.
The core configuration uses \emph{show-when-supported}: it delivers relevant views that the renderer supports and current memory can faithfully express.
Alternative policies additionally require an agent request or a registered issue involving state conditions, scope, or relations across states.
All policies preserve unresolved quantities instead of filling them in to complete a diagram.

An unresolved semantic issue $u$ is eligible for continuation only if
\begin{equation}
\label{eq:repair_eligibility}
\begin{aligned}
g(u)=\mathbf{1}\bigl[
    &\operatorname{AnswerRelevant}(u) \\
    &\land\operatorname{Actionable}(u) \\
    &\land\operatorname{BudgetAvailable}
\bigr].
\end{aligned}
\end{equation}
Eligible issues include ambiguous source readings, revisions lacking source reinspection, invalidated answer dependencies, and conflicting or missing conditions in key states.
An issue must affect the answer and admit a concrete inspection or memory update.
An unknown marker or an ordinary new derivation alone does not trigger semantic repair.
The agent may request continuation, but the harness determines its eligibility.

Protocol recovery responds to a failed output contract and may regenerate an answer with changed physical content.
The labels S and P identify the initiating cause of continuation: semantic repair or protocol recovery.
Both paths share at most one additional model call.
Before that call, the harness delivers current accepted memory, relevant check or protocol feedback, and supported evidence and views.
The resulting submission undergoes the same checks, and residual blocking issues remain \texttt{unresolved} at termination.
Candidates, trigger causes, and statuses are archived, operational retries and their costs are recorded separately.

\begin{algorithm}[htbp]
\caption{PhysScope with a shared continuation budget}
\label{alg:physscope}
\small
\begin{algorithmic}[1]
\Require Statement $Q$, diagram $I$, VLM $\mathcal{L}$, follow-up flag $b\in\{0,1\}$
\Ensure Answer $\hat{y}$ and validation status $z$
\State $\widetilde{\mathcal{M}}
    \gets\mathcal{L}(Q,I,\mathrm{draft})$
\State $\mathcal{M}
    \gets\operatorname{Compile}(\widetilde{\mathcal{M}})$
\State $(\mathcal{M},\mathcal{O}_2)
    \gets\operatorname{PrepareDeliver}(Q,I,\mathcal{M})$
\State $r_2
    \gets\mathcal{L}(Q,I,\mathcal{O}_2,\mathrm{solve})$
\State $(\mathcal{M},\hat{y}_2,z_2,\mathcal{U}_2,p_2)
    \gets\operatorname{Assess}(\mathcal{M},r_2,\mathcal{O}_2)$
\State Archive the two-call answer, status, and memory
\State $(\hat{y},z)\gets(\hat{y}_2,z_2)$
\If{$b=1$ and
    $\bigl(p_2=1\ \lor\
    \exists u\in\mathcal{U}_2:g(u)=1\bigr)$}
    \State Record the initiating cause S/P
    \State Consume the shared continuation budget
    \State $(\mathcal{M},\mathcal{O}_3)
        \gets\operatorname{RefreshDeliver}
        (\mathcal{M},\mathcal{U}_2,p_2)$
    \State $r_3
        \gets\mathcal{L}
        (Q,I,\mathcal{O}_3,\mathrm{follow\mbox{-}up})$
    \State $(\mathcal{M},\hat{y}_3,z_3,\mathcal{U}_3,p_3)
        \gets\operatorname{Assess}
        (\mathcal{M},r_3,\mathcal{O}_3)$
    \State $(\hat{y},z)\gets(\hat{y}_3,z_3)$
\EndIf
\State \Return $(\hat{y},z)$
\end{algorithmic}
\vspace{3pt}
\begin{minipage}{\linewidth}
\footnotesize
\textsc{Assess} checks the output contract
($p=1$ indicates a recoverable protocol failure)
and applies \textsc{UpdateCheck} to a valid submission.
\textsc{UpdateCheck} commits accepted changes,
invalidates registered dependents, and reports unresolved issues.
Delivery operations refresh permitted inputs while preserving history.
The algorithm shows post-core continuation, transport retries
are accounted for separately.
Validation status reports the outcome of harness checks,
not answer correctness.
\end{minipage}
\end{algorithm}
\section{Experiments}
\label{sec:experiments}

We evaluate whether PhysScope improves a VLM agent's physical reasoning and which harness components contribute to these gains.

\subsection{Experimental Setup}
\label{sec:experimental_setup}

\noindent\textbf{Datasets and metrics.}
We evaluate on all 986 base problems in PhysAlign~\citep{liang2026physalignbenchmarkevidencegroundedrole} and an external subset of 1,024 multimodal problems from PhysUniBench~\citep{wang2025physunibench}.
Visual and repair comparisons use all 986 problems. Component ablations and premise interventions use fixed subsets of 512 and 256, respectively.
With the original inputs, deterministic scoring measures content recognition (CAcc), role grounding ($\mathrm{GAcc}_{L}$), and joint success (JAcc).
On matched probes for each backbone, $\mathrm{GAcc}_{\mathrm{Pm}}$ averages grounding with and without verified readings. Both conditions withhold the target role (Appendix~\ref{sec:appendix_scoring}).
Solve runs independently of the probes, without reference answers or ground-truth roles. SolveAcc averages normalized problem credit on PhysAlign. PhysUniBench uses binary final-answer accuracy.
Scores are percentages and differences are in percentage points (pp).

\begin{table}[htbp]
\centering
\scriptsize
\caption{\textbf{Cross-backbone results on PhysAlign~\citep{liang2026physalignbenchmarkevidencegroundedrole}} (986-problem pool). All scores are percentages ($\uparrow$). $G_L$, $G_{\mathrm{Pm}}$, and Solve abbreviate $\mathrm{GAcc}_L$, $\mathrm{GAcc}_{\mathrm{Pm}}$, and SolveAcc. Bold marks the best score within each backbone, including ties.}

\label{tab:physcope_main_results}
\begingroup
\fontsize{9}{10.4}\selectfont
\setlength{\tabcolsep}{1pt}
\renewcommand{\arraystretch}{1.03}
\begin{tabularx}{\linewidth}{@{}l*{5}{>{\raggedleft\arraybackslash}X}@{\hspace{10pt}}*{5}{>{\raggedleft\arraybackslash}X}@{\hspace{10pt}}*{5}{>{\raggedleft\arraybackslash}X}@{}}
\toprule
 & \multicolumn{5}{c}{\textbf{Gemini-3.8-flash~\citep{googledeepmind2026gemini38flash}}} & \multicolumn{5}{c}{\textbf{GPT-5.6-sol~\citep{openai2026gpt56sol}}} & \multicolumn{5}{c}{\textbf{DeepSeek-V4.1-Flash-1017~\citep{deepseekai2026deepseekv41flash}}} \\
\cmidrule(lr){2-6}\cmidrule(lr){7-11}\cmidrule(lr){12-16}
Condition & CAcc & $G_L$ & JAcc & $G_{\mathrm{Pm}}$ & Solve & CAcc & $G_L$ & JAcc & $G_{\mathrm{Pm}}$ & Solve & CAcc & $G_L$ & JAcc & $G_{\mathrm{Pm}}$ & Solve \\
\addlinespace[1.2pt]
B0 & 79.41 & 74.02 & 66.18 & 86.88 & 21.31 & 87.24 & 79.06 & 74.02 & 81.90 & 71.08 & 76.96 & 67.65 & 65.20 & 84.02 & 69.85 \\
B1 & 82.35 & 77.94 & 69.12 & 87.21 & 21.49 & 87.24 & 80.01 & 74.92 & 82.33 & 71.08 & 76.96 & 67.65 & 65.20 & 85.11 & 71.57 \\
H-T & 83.82 & 78.17 & 70.59 & 88.50 & 21.53 & 89.37 & 82.20 & 76.78 & 82.93 & 72.75 & 78.43 & 69.12 & 66.67 & 87.05 & 87.91 \\
H-F@2 & 88.24 & 79.41 & 72.06 & 88.61 & 22.92 & 91.78 & 86.76 & 78.76 & 85.03 & 87.70 & \textbf{96.08} & 91.01 & 88.02 & 95.04 & 89.62 \\
\rowcolor{PhysScopeBlueTint}
H-F+R & \textbf{89.71} & \textbf{83.82} & \textbf{76.47} & \textbf{89.10} & \textbf{30.70} & \textbf{93.25} & \textbf{91.18} & \textbf{84.43} & \textbf{88.68} & \textbf{88.03} & \textbf{96.08} & \textbf{92.65} & \textbf{88.71} & \textbf{95.99} & \textbf{90.94} \\
\addlinespace[3pt]
 & \multicolumn{5}{c}{\textbf{Qwen3.5-9B~\citep{qwen3.5}}} & \multicolumn{5}{c}{\textbf{Qwen3.5-27B~\citep{qwen3.5}}} & \multicolumn{5}{c}{\textbf{InternVL3.5-14B~\citep{wang2025internvl3_5}}} \\
\cmidrule(lr){2-6}\cmidrule(lr){7-11}\cmidrule(lr){12-16}
B0 & 65.81 & 54.55 & 37.55 & 56.53 & 12.12 & 84.14 & 80.29 & 68.12 & 82.01 & 18.39 & 39.58 & 45.83 & 29.17 & 52.68 & 12.71 \\
B1 & 65.81 & 54.55 & 37.55 & 56.53 & 13.52 & 84.21 & 80.31 & 68.28 & 82.17 & 18.41 & 39.58 & 45.86 & 29.17 & 52.68 & 13.01 \\
H-T & 66.50 & 55.00 & 38.33 & 57.00 & 13.60 & 84.42 & 81.00 & 68.67 & 83.00 & 20.71 & 39.58 & 45.86 & 29.17 & 53.57 & 13.01 \\
H-F@2 & 66.74 & 57.08 & 39.67 & 64.11 & 20.73 & 85.33 & 81.75 & 69.25 & 83.93 & 22.01 & 43.75 & \textbf{47.92} & 33.33 & 54.82 & 13.01 \\
\rowcolor{PhysScopeBlueTint}
H-F+R & \textbf{67.17} & \textbf{58.33} & \textbf{40.86} & \textbf{68.93} & \textbf{23.39} & \textbf{85.79} & \textbf{81.83} & \textbf{70.33} & \textbf{83.99} & \textbf{24.54} & \textbf{45.83} & \textbf{47.92} & \textbf{35.42} & \textbf{54.88} & \textbf{14.54} \\
\bottomrule
\end{tabularx}

\endgroup
\end{table}

\noindent\textbf{Backbones and configurations.}
We use Qwen3.5-9B and Qwen3.5-27B~\citep{qwen3.5}, InternVL3.5-14B~\citep{wang2025internvl3_5}, Gemini-3.8-flash~\citep{googledeepmind2026gemini38flash}, GPT-5.6-sol~\citep{openai2026gpt56sol}, and DeepSeek-V4.1-Flash-1017~\citep{deepseekai2026deepseekv41flash}, without updating model weights.
We test five configurations.
\textbf{B0} answers directly. \textbf{B1} adds an ordinary review.
\textbf{H-T} uses two calls with textual memory, source evidence, applicability conditions, and check feedback.
\textbf{H-F@2} adds supported views to Call~2 of H-T.
\textbf{H-F+R} also permits one follow-up for semantic repair or protocol recovery.
Operational retries are counted in resource use. Unless specified, PhysScope denotes H-F+R.
Mechanism studies, the Canvas-CoT comparison~\citep{sun2026canvas}, and external evaluation use Gemini-3.8-flash~\citep{googledeepmind2026gemini38flash}. Update and repair diagnostics also include DeepSeek and GPT.
Appendices~\ref{sec:appendix_controls} and~\ref{sec:appendix_experiments} detail these protocols and results.

\noindent\textbf{Evaluation and reporting.}
We use matched controls and report point estimates. Missing or unaccepted answers receive zero credit.
On PhysAlign, choice and fill-in answers receive binary credit. Calculation and short-answer responses can earn partial credit for explicit derivation steps.
GPT-6 Astra~\citep{openai2026gpt6astra} handles judgments requiring semantic interpretation under rubrics fixed before grading.
An independent audit covers 10\% of samples, and human reviewers resolve all disputed scores (Appendix~\ref{sec:appendix_archive_protocol}).
Appendices~\ref{sec:appendix_candidate_scores} and~\ref{sec:appendix_inference_cost} report raw candidate scores and resource use.

\subsection{Overall Effectiveness}
\label{sec:overall_effectiveness}

\begin{table}[htbp]
\centering
\scriptsize
\caption{\textbf{Visual baseline and external evaluation.} All conditions use Gemini-3.8-flash~\citep{googledeepmind2026gemini38flash}. Scores are percentages ($\uparrow$), bold marks the best in each panel. The external evaluation uses 1,024 problems.}

\label{tab:physcope_baseline_external}
\begingroup
\fontsize{8.5}{10}\selectfont
\setlength{\tabcolsep}{1.5pt}
\renewcommand{\arraystretch}{1}
\begin{tabularx}{\linewidth}{@{}l*{5}{>{\raggedleft\arraybackslash}X}@{}}
\toprule
\multicolumn{6}{@{}l}{\textbf{(a) PhysAlign~\citep{liang2026physalignbenchmarkevidencegroundedrole}}} \\
\addlinespace[1pt]
Method & CAcc & $G_L$ & JAcc & $G_{\mathrm{Pm}}$ & Solve \\
\midrule
B0 & 79.41 & 74.02 & 66.18 & 86.88 & 21.31 \\
Canvas-CoT & 55.64 & 52.21 & 46.81 & 62.11 & 21.36 \\
\rowcolor{PhysScopeBlueTint}
PhysScope & \textbf{89.71} & \textbf{83.82} & \textbf{76.47} & \textbf{89.10} & \textbf{30.70} \\
\bottomrule
\end{tabularx}
\par\vspace{5pt}
\setlength{\tabcolsep}{3.5pt}
\begin{tabularx}{\linewidth}{@{}Xrr@{}}
\multicolumn{3}{@{}l}{\textbf{(b) PhysUniBench~\citep{wang2025physunibench}}} \\
\addlinespace[1pt]
Method & Correct / Total & Solve $\uparrow$ \\
\midrule
B0 & 384 / 1,024 & 37.50 \\
\rowcolor{PhysScopeBlueTint}
PhysScope & 534 / 1,024 & \textbf{52.15} \\
\bottomrule
\end{tabularx}

\endgroup
\end{table}

\noindent\textbf{Results across backbones.}
PhysScope attains the highest scores on all five metrics for every backbone, including ties (Table~\ref{tab:physcope_main_results}).
SolveAcc gains over B0 range from 1.83 to 21.09 pp: DeepSeek improves from 69.85\% to 90.94\%, and GPT from 71.08\% to 88.03\%.
Ordinary review changes SolveAcc by at most 1.72 pp.
The gain is smaller for InternVL (12.71\% to 14.54\%), indicating that the harness remains dependent on the underlying agent's capabilities.
DeepSeek gains 18.06 pp from B0 to H-T, whereas GPT gains 14.95 pp
from H-T to H-F@2.
Gemini obtains its largest increment, 7.78 pp, from enabling repair.
We therefore interpret the Gemini mechanism studies in light of its larger gain from repair.

\noindent\textbf{Visual baseline and external evaluation.}
With Gemini, PhysScope exceeds Canvas-CoT by 9.34 pp in SolveAcc (30.70\% vs.\ 21.36\%, Table~\ref{tab:physcope_baseline_external}).
Canvas-CoT~\citep{sun2026canvas} has similar SolveAcc to B0 (21.36\% vs.\ 21.31\%) but
lower JAcc (46.81\% vs.\ 66.18\%), whereas PhysScope improves both.
The pipelines use different output interfaces. Awarding full credit to all 14 failed Canvas-CoT Solve outputs would close at most 1.42~pp of the 9.34~pp gap (Appendix~\ref{sec:appendix_canvas_execution}).
Across 4,327 Solve and Ground inputs with Gemini, PhysScope averaged 2.08 calls and 96.15k input tokens per task. Canvas-CoT averaged 2.39 calls and 36.60k input tokens per task. Calls include failed invocations. The averages count each Solve or Ground query separately across the 986 parent problems (Appendix~\ref{sec:appendix_inference_cost}).
On PhysUniBench~\citep{wang2025physunibench}, accuracy increases from 37.50\% to 52.15\%, corresponding to 150 additional correct answers among 1,024 problems.

\begingroup
\captionsetup[table]{skip=2pt}
\begin{table}[htbp]
\centering
\caption{\textbf{Complete visual-representation results.} Gemini-3.8-flash, 986-problem pool. Scores are percentages ($\uparrow$), bold includes ties.}
\label{tab:physcope_full_visual}
\begingroup
\fontsize{8.5}{10}\selectfont
\setlength{\tabcolsep}{1.5pt}
\renewcommand{\arraystretch}{1.08}
\begin{tabularx}{\linewidth}{@{}l*{5}{>{\raggedleft\arraybackslash}X}@{}}
\toprule
Condition & CAcc & $G_L$ & JAcc & $G_{\mathrm{Pm}}$ & Solve \\
\midrule
H-T & 83.82 & 78.17 & 70.59 & 88.50 & 21.53 \\
H-S & 86.50 & 78.95 & 71.12 & 88.59 & 22.09 \\
\rowcolor{PhysScopeBlueTint}
H-F@2 & \textbf{88.24} & \textbf{79.41} & \textbf{72.06} & \textbf{88.61} & \textbf{22.92} \\
\bottomrule
\end{tabularx}
\endgroup
\end{table}

\endgroup

\begin{figure}[htbp]
    \centering
    \includegraphics[width=0.69\linewidth]{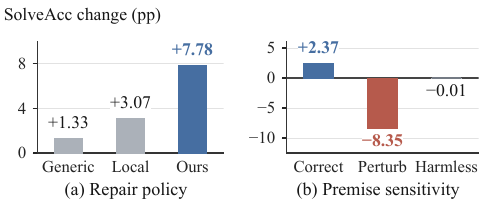}
    \caption{\textbf{Repair and premise sensitivity.}
    SolveAcc changes (pp) with Gemini-3.8-flash~\citep{googledeepmind2026gemini38flash}.
    \textbf{(a)} Review policies versus Stop ($n=986$). Local denotes localized review and Ours denotes controlled repair.
    \textbf{(b)} Premise edits versus predicted memory ($n=256$).
    Axes use separate scales.}
    
    \label{fig:physcope_mechanism_effects}
\end{figure}

\subsection{Mechanism Analysis}
\label{sec:mechanism_analysis}

\paragraph{Visual feedback under matched information.}
\label{sec:value_of_vp}
We compare H-T, H-S (static relation or force views), and H-F@2 (also supporting phase views).
They share the draft from Call~1, memory, source evidence, and check feedback. Every depicted fact, condition, and unknown is also available in their shared text.
H-S uses the same static layout as H-F@2 and leaves its phase region empty. The scores include tasks with unsupported or failed renders.
Static views improve SolveAcc by 0.56 pp, and phase views add 0.83 pp. The corresponding JAcc gains are 0.53 and 0.94 pp (Table~\ref{tab:physcope_full_visual}).
These gains favor visual organization of shared premises. Repeated-run uncertainty is unmeasured.

\paragraph{Scope and dependency management.}
\label{sec:effects_scope_dependency}
\label{sec:effects_scope_denpendency}
Removing scope management reduces SolveAcc from 41.15\% to 24.31\%, while CAcc remains 93.75\% (Table~\ref{tab:physcope_ablations}).
Thus, recognizing content correctly does not suffice to use it under the right physical conditions.
Removing dependency invalidation costs 5.44 pp in SolveAcc, with a 0.02 pp change in CAcc. Without invalidation, registered equations and views can remain available even after a premise they depend on has been replaced.
Removing source checks within Source-Grounded Memory Revision
reduces CAcc by 7.68 pp and SolveAcc by 2.71 pp.
Removing physical checks preserves all four reading/grounding
metrics while reducing SolveAcc by 3.08 pp
(Appendix~\ref{sec:appendix_component_ablations}).
These distinct metric profiles further separate accurate source
reading from maintaining usable physical premises during solving.

\paragraph{Delivery of revised premises.}
We traced premise replacements accepted after Call~2 across Solve and Ground tasks (Appendix~\ref{sec:appendix_activation}).
The harness invalidated at least one registered dependent and delivered the updated snapshot to Call~3 in 216 of 217 such trajectories for Gemini, 187 of 201 for DeepSeek, and 209 of 229 for GPT.
For Gemini, 332 of 667 trajectories with accepted additions delivered new facts to a later call.
Attributing the 5.44~pp loss to delivery requires trajectory-matched ablation results.

\begin{table}[htbp]
\centering
\scriptsize
\caption{\textbf{Mechanism ablations} (512-problem pool). Accuracy is in percent ($\uparrow$), $\Delta$Solve is the change from Full (pp). Bold includes ties. The supplement reports all five metrics.}

\label{tab:physcope_ablations}
\begingroup
\fontsize{8.5}{10}\selectfont
\setlength{\tabcolsep}{1.5pt}
\renewcommand{\arraystretch}{1}
\begin{tabularx}{\linewidth}{@{}>{\raggedright\arraybackslash}p{0.46\linewidth}*{4}{>{\raggedleft\arraybackslash}X}@{}}
\toprule
Configuration & CAcc & JAcc & Solve & $\Delta$Solve \\
\midrule
\rowcolor{PhysScopeBlueTint}
PhysScope (Full) & \textbf{93.75} & \textbf{83.89} & \textbf{41.15} & 0.00 \\
w/o Scope Management & \textbf{93.75} & 74.09 & 24.31 & $-16.84$ \\
w/o Revision Source Check & 86.07 & 80.97 & 38.44 & $-2.71$ \\
w/o Dependency Invalidation & 93.73 & 82.91 & 35.71 & $-5.44$ \\
w/o Physical Checks & \textbf{93.75} & \textbf{83.89} & 38.07 & $-3.08$ \\
\bottomrule
\end{tabularx}

\endgroup
\end{table}

\paragraph{Controlled repair.}
\label{sec:controlled_repair}
A separate policy experiment on all 986 PhysAlign~\citep{liang2026physalignbenchmarkevidencegroundedrole} problems compares follow-up strategies after the H-F@2 core.
Generic review, localized review, and controlled repair reach 24.25\%, 25.99\%, and 30.70\% SolveAcc, respectively, compared with 22.92\% for Stop (Fig.~\ref{fig:physcope_mechanism_effects}a).
Scores retain every parent, including cases without follow-up.
The 4.71 pp advantage over localized review measures the combined policy's net accepted-score gain.
Across 889 pairs of Gemini answers (H-F@2 vs.\ H-F+R, including untriggered problems), 85 gained full content credit and none lost it (Appendix~\ref{sec:appendix_candidate_scores}). These scores measure content independently of acceptance. The 4.71~pp gap uses a different comparator, localized review.

\paragraph{Interventions on physical premises.}
\label{sec:physical_premises}
On 256 problems with verifiable key relations, correcting a memory premise raises SolveAcc from 31.92\% to 34.29\%, perturbing it lowers accuracy to 23.57\%, and a harmless edit leaves it at 31.91\% (Fig.~\ref{fig:physcope_mechanism_effects}b).
These results show sensitivity to premise edits, but correcting one premise need not resolve other reasoning errors.
Corrected premises are used only in this analysis and excluded from the main method scores.

\section{Discussion and Conclusion}
\label{sec:conclusion}

PhysScope explicitly maintains premise applicability and revision consequences in VLM working memory.
Results across six VLMs on PhysAlign~\citep{liang2026physalignbenchmarkevidencegroundedrole} and Gemini on the evaluated PhysUniBench~\citep{wang2025physunibench} subset support this design.
Component studies distinguish source reading, scope enforcement, and dependency maintenance.
These gains reflect complete inference policies. Protocol-triggered regeneration may alter answers, and accepted submissions may still be physically incorrect.
Checks cover declared premises and dependencies under supported rules, but incorrect readings, omitted dependencies, and unsupported arguments may escape them.
The point estimates establish neither repeated-run reliability nor equal inference cost.
Future work should extend check coverage and assess repair reliability across tasks and runs.


\section*{Supplementary Material}
\addcontentsline{toc}{section}{Supplementary Material}
\appendix

\section{Implementation Details}
\label{sec:appendix_harness}

\begin{figure}[htbp]
    \centering
    \includegraphics[width=\textwidth]{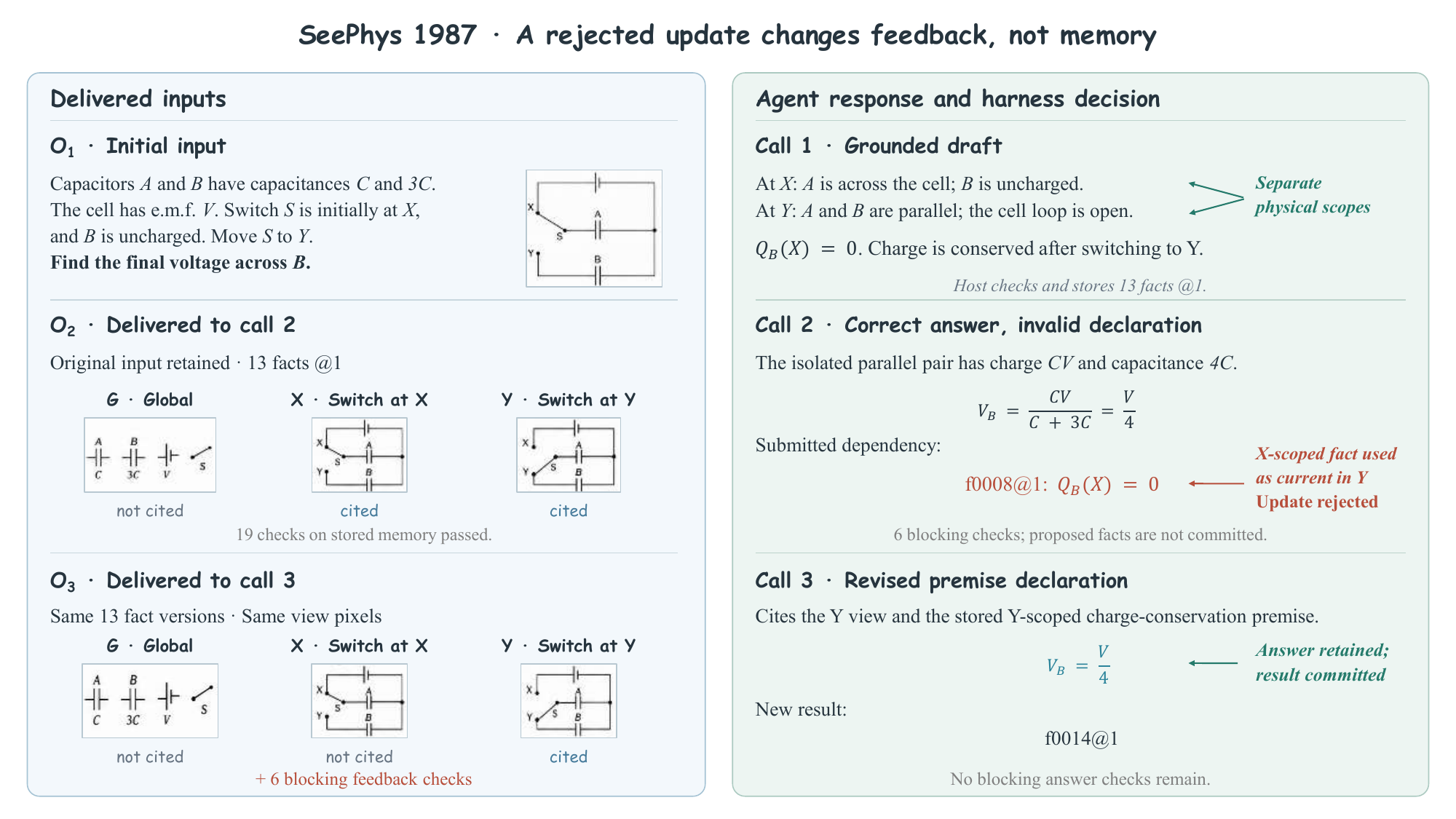}
    \caption{\textbf{Rejected update: feedback changes while accepted memory is preserved.}
    Archived PhysScope execution on SeePhys item~1987.
    Rows pair each input with its response and harness decision.
    Call~2 obtains the correct $V_B=V/4$ but declares X-scoped premises as current inputs in Y.
    Rejection preserves the thirteen fact versions and view pixels in $\mathcal{O}_3$, with six additional blocking feedback entries.
    Call~3 revises the declaration and commits the same answer as \texttt{f0014@1}.
    Text is condensed and diagrams are redrawn from the archived execution, citation labels denote explicit view references.}
    \label{fig:app_rejected_update}
\end{figure}

This appendix details the memory, delivery, and update operations in Algorithm~\ref{alg:physscope}.
Two archived PhysScope executions show how a rejected submission and an accepted new result produce different inputs to the next call (Figs.~\ref{fig:app_rejected_update} and~\ref{fig:app_accepted_update}).
In these traces, Call~1 (\emph{Draft}) prepares premises, Call~2 (\emph{Solve \& Decision-making}) uses the first compiled snapshot to answer and propose updates, and Call~3 is the optional follow-up.

\subsection{Memory Schema and Scope Semantics}
\label{sec:appendix_memory_schema}

Each fact stores an \texttt{id}, \texttt{version}, \texttt{kind}, claim text, and optional value/unit fields.
Entity references (\texttt{owner}, \texttt{target}) identify participants, \texttt{scope}, \texttt{conditions}, \texttt{sources}, and \texttt{depends} record applicability, prerequisites, provenance, and direct dependencies.
References use exact handles, such as \texttt{f0008@1}.
States also carry versions, conditions, and certainty labels, and derived records retain their registered state context.
Only an explicit \texttt{global} scope grants shared applicability.

Claim status, validity, and validation have different meanings.
Claim statuses are \texttt{observed}, \texttt{derived}, \texttt{hypothesis}, and \texttt{unknown}. \texttt{Observed} means a model reading.
Validity tracks usability after updates.
Check outcomes (\texttt{pass}, \texttt{fail}, \texttt{unknown}, \texttt{unsupported}), transaction decisions (accepted, pending, rejected), and answer status are separate fields.

In Case~A, \texttt{f0008@1} reads ``$B$ is initially uncharged'' from \texttt{text:question}, with owner $B$, value $0\,\mathrm{C}$, and scope X.
The derived \texttt{f0013@1} records charge conservation in Y, depending on the switch condition \texttt{f0010@1} and parallel-connection relation \texttt{f0012@1}.
Initial charge can legitimately support a conservation argument, but its role as an initial condition must remain distinct from a current charge in Y.

\paragraph{Operational applicability rule.}
For an ordinary, untyped fact $f$ used directly in state $s$, the domain check is
\begin{equation}
 \operatorname{Applies}(f,s)=
 \mathbf{1}[\texttt{global}\in\operatorname{scope}(f)
 \;\lor\;s\in\operatorname{scope}(f)].
 \label{eq:app_scope_contract}
\end{equation}
A local derivation checks each input in its declared producer context, its output must belong to that context.
Historical ancestors retain their producer contexts and exact versions.
Thus, preserving a historical record and using it as a current value are distinct operations.

\paragraph{Initial parameters and current values.}
A fixed parameter $v_0:=v(A)$ may be declared globally when its meaning remains the initial velocity throughout the argument, this does not assert $v(B)=v_0$.
By contrast, a current-angle reading $\theta(A)$ belongs to A, and $\theta(B)$ belongs to B.
The following schema example uses the same applicability rule as Case~A:
\begin{center}
\small
\setlength{\tabcolsep}{3pt}
\begin{tabular}{@{}lll@{}}
\toprule
Record & Declared scope & Direct use at B \\
\midrule
$v_0:=v(A)=0$ & global & Allowed as declared \\
$\theta(A)=30^\circ$ & A & Blocked \\
$\theta(B)=53^\circ$ & B & Allowed as declared \\
\bottomrule
\end{tabular}
\end{center}
At B, a derivation that uses $v_0$ as an initial parameter and $\theta(B)$ as its current angle satisfies this domain contract.
Substituting the A-scoped angle fails the contract.
In Case~A, the rejected derivation explicitly treated X-scoped inputs as current in Y, the accepted continuation instead cites the stored Y-scoped conservation relation.
The numerical result stays $V/4$, so the trace isolates a change in declared support.

\paragraph{Typed temporal contracts.}
For typed quantities, \texttt{value\_at} distinguishes invariant, interval, and point values, with an explicit state and boundary side.
An initial-role input must match the receiving interval's start boundary on its permitted side, the same point value cannot be used as an interval-current input.
The implemented continuity transfer checks adjacent states, quantity and owner identity, frame and unit agreement, matching values, and an explicit continuity condition.
Ordinary untyped records use Eq.~\eqref{eq:app_scope_contract}, merely naming an input role \texttt{initial} does not override it.
These checks enforce supplied declarations.
An incorrect \texttt{global} annotation can therefore satisfy the domain test, and a physically valid argument still requires correct premises and sufficient declared support.

State evolution and premise correction have different semantics.
Recording a later velocity adds a state-specific fact while preserving the initial velocity, correcting an erroneous reading replaces a version and invalidates its registered dependents.
Neither a change of physical state nor the presence of an older record alone warrants invalidation.

\subsection{Source Inspection and Delivered Snapshots}
\label{sec:appendix_snapshots}

\paragraph{Source access and delivery.}
\texttt{inspect\_source} takes an immutable source identifier and a text span or image box.
Text spans use zero-based, half-open character offsets. Image boxes use normalized $(x_0,y_0,x_1,y_1)$ coordinates from the upper-left origin, rounded outward to pixel boundaries.
Returned evidence contains the localized text or crop, source coordinates, identifier, and delivery record.
Original inputs remain in the conversation history, neither case requests an additional inspection.

A delivered snapshot $\mathcal{O}_t$ fixes the memory versions, source evidence, check feedback, and views supplied to call~$t$.
Memory and checks are serialized as text, visual evidence is attached as image content.
Receipts associate view identifiers and image hashes with the receiving request, including retained history.
Rendering alone does not establish delivery.
In the figures, $\mathcal{O}_1$ denotes the initial question, image, and task contract, before memory construction, $\mathcal{O}_2$ is the first compiled snapshot.
View citation labels report explicit \texttt{answer\_uses} declarations, not measured attention.

\paragraph{Case A: feedback without a memory change.}
After Call~1, $\mathcal{O}_2$ contains thirteen facts at version~1, nineteen passing memory checks, and global, X, and Y views.
Call~2 returns $V_B=CV/(C+3C)=V/4$ but submits a Y-scoped derivation with X-scoped inputs declared as current, including \texttt{f0008@1}.
The six blocking checks include domain violations and missing support caused by rejection, they are not six independent physical mistakes.
$\mathcal{O}_3$ therefore retains the fact versions and view pixels while adding feedback.
Call~3 cites Y and its stored charge-conservation premise, then commits \texttt{f0014@1} with the unchanged value $V/4$.
Repair corrects the declaration without changing the numerical answer or visual content.

\begin{figure}[htbp]
    \centering
    \includegraphics[width=\textwidth]{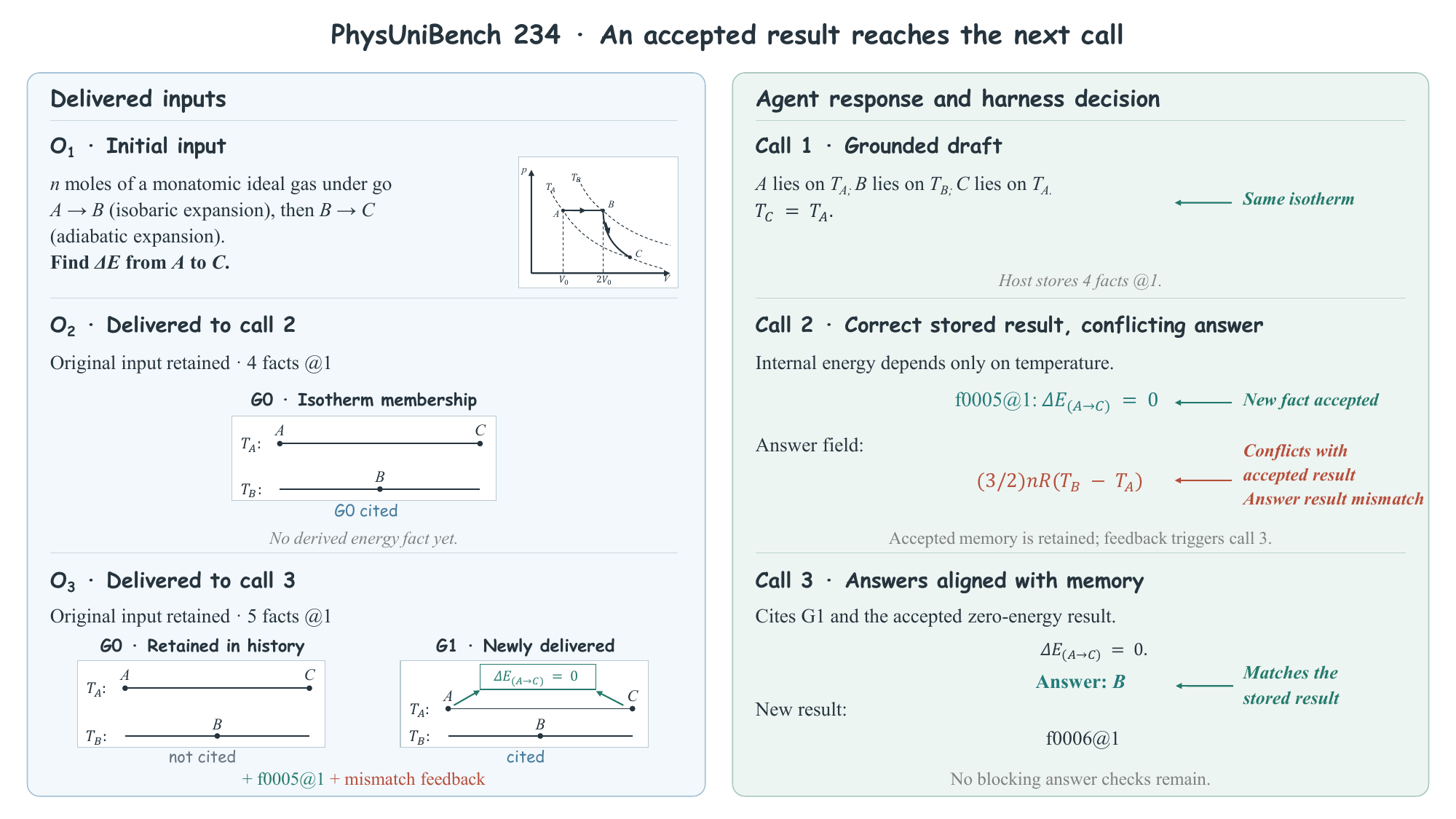}
    \caption{\textbf{Accepted update: a new result is delivered before answer repair.}
    Archived PhysScope execution on PhysUniBench item~234.
    Call~2 commits \texttt{f0005@1}, recording $\Delta E_{A\to C}=0$, but its answer field contains $\frac{3}{2}nR(T_B-T_A)$.
    $\mathcal{O}_3$ retains that fact among five records and delivers G1 with mismatch feedback, G0 remains in history.
    Call~3 cites G1 and returns choice~B, supported by the accepted zero-energy result.
    Text is condensed and diagrams are redrawn from the archived execution, G0 and G1 summarize isotherm membership.}
    \label{fig:app_accepted_update}
\end{figure}

\paragraph{Case B: a memory change followed by answer repair.}
Here, $\mathcal{O}_2$ contains four facts at version~1 and view G0, which records that A and C lie on the same isotherm.
Call~2 adds \texttt{f0005@1}, the derived result $\Delta E_{A\to C}=0$, while its answer field gives $\frac{3}{2}nR(T_B-T_A)$.
The fact transaction is accepted, but the subsequent answer check reports \texttt{answer\_result\_mismatch}.
The next request exposes five facts, mismatch feedback, and G1 with the new result, G0 remains in history.
Call~3 cites G1 and adds \texttt{f0006@1}, mapping the zero-energy result to choice~B.
The original four facts remain at version~1: this is an addition, not a replacement.
Acceptance changes memory, subsequent delivery exposes that change to the agent.
The trace does not isolate G1's effect from the accompanying text and feedback.

\subsection{Rendering Constraints and Check Coverage}
\label{sec:appendix_rendering_checks}

The renderer supports global and local relation views and comparisons between declared states.
Temporal or mode panels require a meaningful state difference, a single state uses a local view.
Unknowns and conditional states retain their labels, and missing defining conditions prevent a definite depiction.
Schematic layout adds no physical premises.
Accordingly, Case~A compares switch modes, while Case~B uses relational views of isotherm membership, G0 and G1 denote successive memory views, not different thermodynamic states.

Checks operate on declared records and delivery receipts.
They cover reference and graph integrity, exact versions, input/output scopes, dependency validity, source reinspection, and cited-view scope and delivery.
Answer checks require accepted support and agreement between answer slots and their declared results.
Each check records its rule, affected records, outcome, enforcement status, and repairability.
Unknown or unsupported outcomes block an answer only under the applicable enforcement and relevance rules.

Both cases use the ordinary relational profile, which indexes derivations as \texttt{declared\_only}.
Their domain and answer checks establish declared consistency, without proving the charge-conservation or ideal-gas arguments.
Finite physical templates, when enabled, require specified quantities, units, conditions, and domains, they do not constitute a general physics solver.
Ordinary scope checks, typed temporal contracts, and finite physical rules therefore have distinct coverage.
A declared dependency can be structurally valid without its physical derivation being verified.
The physical-check ablation measures the enabled module's contribution within its evaluation configuration, it does not measure universal rule coverage or per-check detection accuracy.

\subsection{Revision Commit and Dependency Invalidation}
\label{sec:appendix_revision_details}

\paragraph{Staging and commit.}
\textsc{UpdateCheck} binds a submission to its delivered snapshot and stages changes in a copy of memory.
It resolves local keys and validates references, scopes, and the dependency graph before committing the core transaction.
A source-backed replacement also requires matching reinspection evidence delivered to the proposing call.
Missing reinspection leaves the transaction pending, a violated update contract rejects it.
Both preserve active facts and retain diagnostic feedback.
Accepted additions receive new identifiers at version~1, replacements increment the existing identifier's version.

\paragraph{Propagation and answer validation.}
Replacement invalidates the old premise and its registered descendants, unrelated records remain usable.
For a registered chain $f_i^{(v)}\to d\to q$, where $d$ is a derivation and $q$ a dependent result or view, committing $f_i^{(v+1)}$ requires $d$ and $q$ to be re-established against current premises.
References are not redirected, and invalidation does not compute new conclusions.
Neither displayed case instantiates this replacement rule: Case~A rejects an extension, while Case~B adds records.
Older artifacts can remain visible in the preserved history, but invalidated versions cannot serve as active support in a new accepted submission.
This restriction governs the explicit workspace and submission checks, it cannot guarantee that the VLM ignores stale information in its internal reasoning.

Answer validation follows the memory decision.
It detects unavailable support after rejection in Case~A and a mismatch after acceptance in Case~B, preserving independently accepted memory updates.
Coverage is limited to explicit declarations: omitted dependencies cannot be invalidated, and removing a flagged dependency does not establish a complete physical argument.

\subsection{Tool Scheduling and Termination}
\label{sec:appendix_scheduling}

The core \emph{show-when-supported} policy delivers views that current memory can faithfully express.
Semantic repair requires an enforced, answer-relevant blocking issue, an available corrective action, and remaining call budget.
Actions include source reinspection and correction of memory or answer-support declarations.
An unknown marker, a new fact, or an agent request alone is insufficient.
Both cases trigger declaration repair without a new source crop.

Semantic repair and protocol recovery share one additional call after the two-call core.
The harness refreshes memory, evidence, checks, and supported views before that call while retaining history.
Updates affect only later inputs, never the snapshot that produced them.
At budget exhaustion, remaining blocking checks accompany the latest candidate as \texttt{unresolved}, both the two-call answer and repair outcome are preserved.
Termination does not certify the candidate or resolve the remaining uncertainty.
Undelivered views remain audit artifacts, semantic repair, protocol recovery, and operational retries retain separate cost records.
Protocol recovery can request a newly generated answer, allowing changes to physical content as well as submission form.
The S/P categories in Table~\ref{tab:app_repair_outcomes} identify the recorded trigger cause, independently of the resulting answer changes.
Neither path grants another continuation after the shared allowance is spent.
\textsc{Assess} in Algorithm~\ref{alg:physscope} separates output-contract validation from \textsc{UpdateCheck}, an invalid submission cannot silently commit unvalidated changes.

\section{Experimental Protocols}
\label{sec:appendix_protocols}
\label{sec:appendix_experiment_protocol}

This appendix specifies the evaluation pools, scoring rules, and comparison protocols.
Implementation details are given in Appendix~\ref{sec:appendix_harness}, and complete diagnostic results in Appendix~\ref{sec:appendix_experiments}.

\subsection{Datasets and Evaluation Pools}
\label{sec:appendix_data_pools}

Table~\ref{tab:app_evaluation_pools} summarizes the reported evaluation pools.
The main evaluation uses all 986 parent problems in PhysAlign~\cite{liang2026physalignbenchmarkevidencegroundedrole}, covering its full release rather than only its test partition.
Visual-feedback and repair comparisons reuse this pool, component ablations and premise interventions use subsets of 512 and 256 parents, respectively.
The external evaluation uses 1,024 multimodal physics problems sampled from PhysUniBench~\cite{wang2025physunibench}.
Each comparison uses its own reference configuration on the same pool: the full-method score on 512 problems is not interchangeable with that on 986 problems.
The 512- and 256-parent studies are targeted diagnostic subsets, not additional independent replications of the full-pool score.

\begin{table}[htbp]
    \centering
    \caption{\textbf{Evaluation pools.} Counts refer to parent problems, not localized probes or model calls. ``Gemini'' denotes Gemini-3.8-flash.}
    \label{tab:app_evaluation_pools}
    \footnotesize
    \setlength{\tabcolsep}{3pt}
    \begin{tabular}{@{}llrl@{}}
        \toprule
        Study & Dataset & Parents & Backbones \\
        \midrule
        Main results & PhysAlign & 986 & 6 \\
        Canvas-CoT comparison & PhysAlign & 986 & Gemini \\
        Visual feedback / repair & PhysAlign & 986 & Gemini \\
        Component ablations & PhysAlign & 512 & Gemini \\
        Premise interventions & PhysAlign & 256 & Gemini \\
        External evaluation & PhysUniBench & 1,024 & Gemini \\
        \bottomrule
    \end{tabular}
\end{table}

Local Read/Grounding and original-problem Solve are separate evaluation tasks.
Solving receives the original question and images, without local-probe targets, reference solutions, or ground-truth role assignments.
Verified local readings are supplied only in the designated +GT grounding condition.
Reference-based premise corrections belong exclusively to the diagnostic intervention experiment, they are excluded from ordinary method evaluation.

\subsection{Metrics and Answer Scoring}
\label{sec:appendix_scoring}

\paragraph{Local correctness.}
We follow PhysAlign's deterministic recognition and role-grounding definitions~\cite{liang2026physalignbenchmarkevidencegroundedrole}.
For probe $j$, let $\hat c_j$ and $\hat u_j$ be the predicted reading and candidate identifier, $c_j$ and $r_j$ the reference reading and role, $N_j$ the prescribed normalizer, $E_j$ the candidate set, and $\phi_j$ its role mapping.
The scores are
\begin{equation}
    \begin{aligned}
        C_j &= \mathbf{1}\!\left[N_j(\hat c_j)=N_j(c_j)\right], \\
        G_j &= \mathbf{1}\!\left[\hat u_j\in E_j\ \land\ \phi_j(\hat u_j)=r_j\right], \\
        J_j &= C_jG_j.
    \end{aligned}
    \label{eq:app_local_scores}
\end{equation}
Reading is evaluated at the specified evidence anchor.
The normalizer admits only predefined formatting equivalences, it does not repair a physical interpretation.
The two fields are scored independently, so a correct reading can coexist with an incorrect owner.
No language-model judge is used for these local scores.

\paragraph{Balanced aggregation.}
For each parent pool, let $\mathcal S_{it}$ contain the probes of parent $i$ and task type $t$ in evaluation set $\mathcal S$.
Write $m_{it}^{\mathcal S}=|\mathcal S_{it}|$, let $N_t^{\mathcal S}$ count parents represented in type $t$, and let $\alpha_t^{\mathcal S}\geq0$ be the fixed type weights, summing to one.
PhysAlign's aggregation is
\begin{equation}
    \operatorname{Agg}_{\mathcal S}(z)
    =\sum_t \frac{\alpha_t^{\mathcal S}}{N_t^{\mathcal S}}
      \sum_{i:m_{it}^{\mathcal S}>0}\frac{1}{m_{it}^{\mathcal S}}
      \sum_{j\in\mathcal S_{it}} z_j.
    \label{eq:app_balanced_aggregation}
\end{equation}
Thus, parents with more annotations do not receive more weight within a task type.
On $\mathcal L$, the probes with independent reading and grounding targets, we report
\begin{equation}
    \begin{aligned}
        \mathrm{CAcc} &= 100\operatorname{Agg}_{\mathcal L}(C), \\
        \mathrm{GAcc}_{L} &= 100\operatorname{Agg}_{\mathcal L}(G^{\mathrm{base}}), \\
        \mathrm{JAcc} &= 100\operatorname{Agg}_{\mathcal L}(CG^{\mathrm{base}}).
    \end{aligned}
    \label{eq:app_reported_local_metrics}
\end{equation}
Here $C$ uses the Base condition.
$\mathrm{GAcc}_{L}$ includes both correct and incorrect readings, JAcc measures their joint success on each probe.

\paragraph{Combined Base/+GT grounding.}
Our tables combine the two reading conditions into one $\mathrm{GAcc}_{\mathcal P_m}$ column.
For backbone $m$, $\mathcal P_m\subseteq\mathcal L\cap\mathcal P$ denotes the scheduled matched probe support, where $\mathcal P$ contains probes with valid +GT inputs.
Base and +GT retain the same question, evidence anchors, candidates, and scoring rules, and use separate contexts.
+GT additionally provides verified local readings, with the queried role withheld.
With identical support and weights for both conditions, pooling their records gives
\begin{equation}
    \mathrm{GAcc}_{\mathcal P_m}
    =100\operatorname{Agg}_{\mathcal P_m}
      \!\left(\frac{G^{\mathrm{base}}+G^{+\mathrm{GT}}}{2}\right).
    \label{eq:app_combined_grounding}
\end{equation}
The denominator accounts for both condition records per probe, this is a combined accuracy, not a sum of percentages or a Base-to-+GT gain.
Supplied readings receive no CAcc or JAcc credit.
The combined score should therefore be interpreted as grounding across two input conditions, not as performance without reading assistance.

\paragraph{Original-problem scoring.}
For a pool of $N$ parents, PhysAlign SolveAcc averages normalized problem credit:
\begin{equation}
    \mathrm{SolveAcc}=\frac{100}{N}\sum_{i=1}^{N}s_i,
    \qquad s_i\in[0,1].
    \label{eq:app_solve_score}
\end{equation}
Choice answers use fixed matching rules, fill-in answers use binary reference-equivalence criteria, and short-answer or calculation responses receive rubric-based credit for conclusions and visible derivation steps.
Scores are normalized by the problem's maximum credit, the rubric and judge settings are specified in Appendix~\ref{sec:appendix_archive_protocol}.
Each parent has equal weight.
Consequently, a PhysAlign SolveAcc difference cannot generally be converted into a count of fully solved problems.
The PhysUniBench experiment instead reports binary final-answer correctness, so Eq.~\eqref{eq:app_solve_score} uses $N=1{,}024$ and $s_i\in\{0,1\}$ there.

\paragraph{Missing outputs and validity.}
Evaluation retains the scheduled denominator.
Unanswered API requests, including timeouts, receive zero credit, failed local fields receive zero on their applicable metrics.
This end-to-end failure policy differs from restricting evaluation to returned responses.
Consequently, our API-based B0 scores can differ from benchmark reports that use a different response-availability policy.
Fields without a recognition target, including supplied +GT readings, are inapplicable rather than incorrect.
For solving, system failures and candidates with unresolved blocking dependencies are unsuccessful under the harness protocol, a valid answer produced by the predefined fallback is scored normally.
The two-call candidate and any repaired candidate are retained separately.
Harness acceptance does not itself establish answer correctness.
All reported scores are percentages, and differences are in percentage points.

\subsection{Backbones and Baseline Configurations}
\label{sec:appendix_baselines}

The main evaluation uses Gemini-3.8-flash~\citep{googledeepmind2026gemini38flash}, GPT-5.6-sol~\citep{openai2026gpt56sol}, DeepSeek-V4.1-Flash-1017~\citep{deepseekai2026deepseekv41flash}, Qwen3.5-9B~\citep{qwen3.5}, Qwen3.5-27B~\citep{qwen3.5}, and InternVL3.5-14B~\citep{wang2025internvl3_5}.
Each backbone is evaluated under B0, B1, H-T, H-F@2, and H-F+R without updating its weights.
Controlled mechanism studies, Canvas-CoT~\citep{sun2026canvas}, and external evaluation use Gemini-3.8-flash~\citep{googledeepmind2026gemini38flash}.
Update and repair diagnostics additionally include DeepSeek and GPT.

\paragraph{Direct and reviewed answers.}
B0 answers the original task in one call.
B1 makes a second call that receives the original inputs and first answer and requests a general review.
It controls for an additional opportunity to reconsider the answer.

\paragraph{Harness variants.}
H-T uses a drafting call followed by a solving call with textual memory, source evidence and crops, applicability conditions, unknowns, and check feedback.
It excludes synthesized views but retains original visual evidence.
H-S adds static relation or force views.
H-F@2 additionally supports temporal-phase or mode comparisons.
H-F+R permits at most one further call, shared between semantic repair and protocol recovery.
Semantic eligibility requires an answer-relevant issue with an actionable inspection or update, protocol recovery addresses an output-contract failure.
Problems with neither semantic repair nor protocol recovery retain the two-call result.
Unless qualified, PhysScope denotes H-F+R.
Compilation, source inspection, rendering, and checks use programmatic tools rather than additional model critics.

\paragraph{Comparison scope and budgets.}
The Canvas-CoT~\cite{sun2026canvas} comparison assesses the two complete reasoning pipelines with the same backbone.
Component effects are examined separately through the controlled studies below.
Scheduled model stages are one for B0, two for B1 and the core harness variants, and at most three for H-F+R, including its shared S/P continuation.
Equal call counts do not imply equal token usage, image input, latency, or cost.
Operational request retries are separate from these model stages and included in execution costs.
Protocol recovery can regenerate the answer and change its content, its category records the initiating issue rather than the resulting edit.
Table~\ref{tab:app_repair_outcomes} counts distinct triggered Solve problems and partitions them by semantic or protocol cause, separately from request counts.
When first-call outputs are reused across experimental arms, their cost still belongs to each method's standalone execution, reuse only reduces the experiment's actual bill.

\subsection{Controlled Comparison Protocols}
\label{sec:appendix_controls}

\paragraph{Visual representation.}
H-T, H-S, and H-F@2 share the first-call draft, memory snapshot, source crops, conditions, unknowns, and check feedback.
Every relation or condition shown in an added view also appears in their shared text.
The \emph{supported-fixed} policy delivers relevant views whenever the registered content and renderer support a faithful depiction.
H-S and H-F@2 use identical static content and layout, the phase region is left empty in H-S.
Static spatial-region comparisons belong to both variants, while H-F@2's additional panels represent temporal phases or mode changes.
Missing states, unsupported views, and rendering failures remain in the full evaluation pool.
No unregistered physical premise is introduced to complete a diagram.

For a metric $S$ in percentage units, the representation increments are
\begin{equation}
    \begin{aligned}
        \Delta_{\mathrm{static}} &= S(\text{H-S})-S(\text{H-T}), \\
        \Delta_{\mathrm{phase}} &= S(\text{H-F@2})-S(\text{H-S}).
    \end{aligned}
    \label{eq:app_visual_increments}
\end{equation}
These comparisons test the presentation of shared information.
By contrast, H-T versus B1 changes both task decomposition and workspace access, so its gain cannot be attributed to memory storage alone.

\paragraph{Component ablations.}
The 512-parent study removes one mechanism from the full harness at a time.
Without scope management, stage annotations remain but runtime applicability enforcement is disabled.
Without revision source checks, source reinspection no longer gates changes to original readings.
Without dependency invalidation, records retain their versions and declared dependencies, but upstream revisions no longer invalidate affected descendants.
Without physical checks, the finite physical-rule module is disabled.
An ablated mechanism must not be reinstated through another filtering or feedback path.
Its intervention takes effect before the next solving call, changing only final acceptance would test validation rather than the subsequent reasoning process.
In particular, invalidation can influence reasoning only when a revision affects registered dependencies and a later call receives the changed workspace.

\paragraph{Review and controlled repair.}
This separately executed study compares Stop, Generic review, Localized review, and Controlled repair on all 986 PhysAlign parents.
Stop preserves the H-F@2 result.
Generic review requests reconsideration. Localized review directs attention to an identified issue, Controlled repair applies the source, revision, and dependency checks described in Appendix~\ref{sec:appendix_scheduling}.
Controlled repair requires an answer-relevant ambiguity, a reading revision lacking required source inspection, an invalid answer dependency, or a conflicting or missing state condition, together with an actionable correction.
An unknown quantity alone does not justify another call.
S and P share a one-call continuation budget, problems without follow-up retain their two-call result in the full-pool score.
Reaching the budget preserves residual issues as unresolved, it does not certify the latest candidate.
Views generated only after the final call are audit artifacts, not evidence used by that call.
The trigger counts in Table~\ref{tab:app_repair_outcomes} belong to the main-run diagnostics, not this separate policy experiment.
Sharing the 986-parent evaluation set does not establish shared generated candidates or identical trigger membership across experiments.
We therefore interpret the review results as full-pool policy comparisons, without attributing their differences to a fixed number of semantic-only repairs.

\paragraph{Premise interventions.}
The 256-parent study compares Predicted memory with Corrected premise, Perturbed premise, and Harmless edit conditions.
The first two edits change the physical content of a selected premise, the harmless edit preserves that content and controls for editing alone.
Each condition is compared with Predicted on this pool, with Read/Grounding targets excluded from the Solve task.
Reference-assisted correction is a diagnostic intervention, not an inference-time capability of PhysScope.
The resulting differences measure sensitivity to the edited premises. They do not establish that all unedited memory is correct or that every error is detectable by the harness.

\paragraph{Statistical interpretation.}
We report aggregate point estimates without confidence intervals or repeated-run variance estimates.
Small differences, including the static/phase increments and the harmless-edit contrast, are descriptive rather than claims of statistical significance.
The independent unit for population-level uncertainty is the parent problem, multiple Ground queries and candidate versions from one parent are not independent replications.

\subsection{Scoring and Inference Settings}
\label{sec:appendix_archive_protocol}

\paragraph{Judge and scoring rubric.}
Non-deterministic judgments use GPT-6 Astra~\citep{openai2026gpt6astra} with medium reasoning effort, a 16,384-token output limit, no explicit temperature setting, and a 900-second timeout.
Rubrics are fixed before candidate answers are inspected.
They follow reference point allocations when available, otherwise, requested subparts receive equal weight, distributed equally across independent steps.
Choice answers use fixed matching rules, and fill-in answers receive binary credit for reference equivalence.
Calculation and short-answer responses can also earn credit for correct steps in the submitted explanation.
The same judge resolves equivalence cases that require semantic assessment.
Scoring uses the designated final-answer fields and, where applicable, the visible solution, omitted steps and hidden deliberation receive no credit.
Each score is normalized as $\sum_k\mathrm{awarded}_k/\sum_k\mathrm{maximum}_k$.

\paragraph{Quality control.}
An independent audit covers 10\% of samples.
Human reviewers examine all disputed samples, resolve reference or scoring disagreements, and assign the final scores.
The same frozen rubric applies to each candidate for a given problem.
Appendix~\ref{sec:appendix_judge_prompts} provides the grading instructions.

\paragraph{Inference settings.}
The Gemini solver and Canvas-CoT~\citep{sun2026canvas} critic use Gemini-3.8-flash~\citep{googledeepmind2026gemini38flash} with medium reasoning effort, temperature~1, a 32,768-token output limit, automatic image detail, and a 900-second timeout.
Each method's resource use includes its initial draft, even when that draft is shared across experimental conditions.

\section{Complete Experimental Results}
\label{sec:appendix_experiments}

Controlled mechanism comparisons use Gemini-3.8-flash, update diagnostics also include DeepSeek and GPT.
Visual-feedback and repair comparisons use the 986-problem PhysAlign pool. Component ablations and premise interventions use subsets of 512 and 256 problems, respectively.
Reading/grounding and solving are evaluated independently.
Scores are aggregate point estimates, and changes are reported in percentage points (pp) relative to the corresponding control.

\subsection{Interpreting the Visual-Feedback Results}
\label{sec:appendix_visual_results}

Table~\ref{tab:physcope_full_visual} in the main paper reports all five metrics. From H-T to H-F@2, CAcc rises by 4.42~pp.
All three conditions use the same two scheduled calls, and every depicted premise also appears in the shared text (Appendix~\ref{sec:appendix_controls}).
The observed phase increments are larger than the static increments on JAcc and SolveAcc. Without repeated-run uncertainty estimates, this ordering remains descriptive.

\subsection{Complete Component Ablations}
\label{sec:appendix_component_ablations}

\begin{table}[htbp]
\centering
\caption{\textbf{Complete mechanism ablations.} The selected pool contains 512 problems. All scores are percentages ($\uparrow$), bold includes ties.}
\label{tab:physcope_full_ablation}
\begingroup
\fontsize{9}{10.4}\selectfont
\setlength{\tabcolsep}{1.5pt}
\renewcommand{\arraystretch}{1.08}
\begin{tabularx}{\linewidth}{@{}l*{5}{>{\raggedleft\arraybackslash}X}@{}}
\toprule
Condition & CAcc & $G_L$ & JAcc & $G_{\mathrm{Pm}}$ & Solve \\
\midrule
\rowcolor{PhysScopeBlueTint}
PhysScope (Full) & \textbf{93.75} & \textbf{83.89} & \textbf{83.89} & \textbf{88.84} & \textbf{41.15} \\
w/o Scope Management & \textbf{93.75} & 76.13 & 74.09 & 86.93 & 24.31 \\
w/o Revision Source Check & 86.07 & 82.11 & 80.97 & \textbf{88.84} & 38.44 \\
w/o Dependency Invalidation & 93.73 & 82.99 & 82.91 & \textbf{88.84} & 35.71 \\
w/o Physical Checks & \textbf{93.75} & \textbf{83.89} & \textbf{83.89} & \textbf{88.84} & 38.07 \\
\bottomrule
\end{tabularx}
\endgroup
\end{table}

\begin{figure}[htbp]
    \centering
    \includegraphics[width=\textwidth]{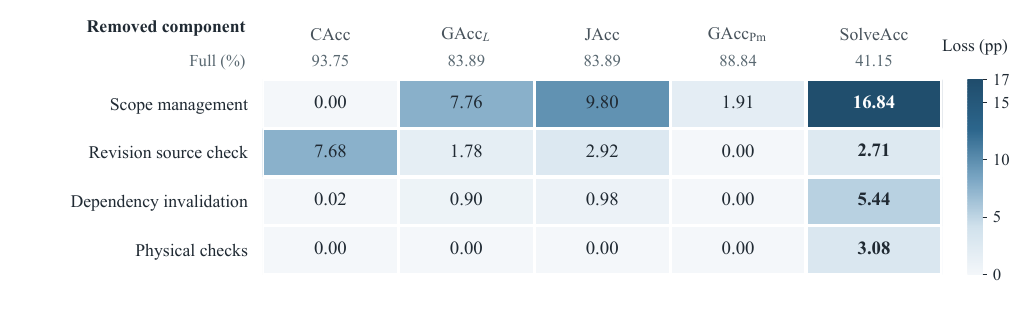}
    \caption{\textbf{Accuracy losses under component removal.}
    Gemini-3.8-flash is evaluated on the 512-problem ablation pool.
    Each cell reports the full-system score minus the score after removing one component, in percentage points.
    All cells share one color scale, full-system scores are shown above the columns.
    Each removal is evaluated with the remaining components enabled.}
    \label{fig:physcope_ablation_losses}
\end{figure}

Scope management has the largest SolveAcc effect among the tested removals (Table~\ref{tab:physcope_full_ablation}, Fig.~\ref{fig:physcope_ablation_losses}).
Without it, CAcc remains 93.75\%, but JAcc falls by 9.80 pp and SolveAcc by 16.84 pp.
Removing dependency invalidation reduces CAcc by only 0.02 pp while reducing SolveAcc by 5.44 pp.
These contrasts support maintaining premise applicability and revision consequences in addition to accurate source recognition.

Source and physical checks have different metric profiles.
Removing revision source checks reduces CAcc by 7.68 pp and SolveAcc by 2.71 pp, consistent with their role in constraining changes to source readings.
Removing physical checks leaves all four reading/grounding metrics unchanged but reduces SolveAcc by 3.08 pp.
A component can therefore improve solving without increasing measured recognition or grounding accuracy.
Each ablation measures a removal within the full system, the losses do not constitute an additive decomposition of its overall gain.
The study also does not estimate performance conditional on a particular check firing: that requires per-trajectory check activations and matched outcomes.

\subsection{Complete Repair Results}
\label{sec:appendix_repair_results}

\begin{table}[htbp]
\centering
\caption{\textbf{Complete repair results.} Separate policy experiment on all 986 PhysAlign parents. Scores are percentages ($\uparrow$), bold includes ties. Trigger counts from the main-run diagnostics are not reused here.}
\label{tab:physcope_full_repair}
\begingroup
\fontsize{8.5}{10}\selectfont
\setlength{\tabcolsep}{1.5pt}
\renewcommand{\arraystretch}{1.08}
\begin{tabularx}{\linewidth}{@{}>{\raggedright\arraybackslash}p{0.32\linewidth}*{5}{>{\raggedleft\arraybackslash}X}@{}}
\toprule
Condition & CAcc & $G_L$ & JAcc & $G_{\mathrm{Pm}}$ & Solve \\
\midrule
Stop & 88.24 & 79.41 & 72.06 & 88.61 & 22.92 \\
Generic review & 88.92 & 82.07 & 74.07 & 88.89 & 24.25 \\
Localized review & 88.92 & 82.55 & 75.11 & 88.99 & 25.99 \\
\rowcolor{PhysScopeBlueTint}
Controlled repair & \textbf{89.71} & \textbf{83.82} & \textbf{76.47} & \textbf{89.10} & \textbf{30.70} \\
\bottomrule
\end{tabularx}
\endgroup
\end{table}

Table~\ref{tab:physcope_full_repair} reports a separate policy experiment on the full 986-parent pool.
Stop ends after the H-F@2 core, the other conditions apply their respective continuation policies and retain all parents in the score.
Generic review, localized review, and controlled repair improve SolveAcc over Stop by 1.33, 3.07, and 7.78 pp, respectively.
Controlled repair also raises JAcc from 72.06\% to 76.47\%.
Its 4.71 pp SolveAcc advantage over localized review measures source checks, explicit memory updates, dependency handling, and submission acceptance jointly.
The candidate transitions in Appendix~\ref{sec:appendix_candidate_scores} compare H-F@2 with H-F+R, not Localized with Controlled. They therefore cannot allocate this 4.71 pp difference between content and acceptance.

The H-F+R gain over H-F@2 varies across backbones, from 0.33 pp for GPT to 7.78 pp for Gemini (Table~\ref{tab:physcope_main_results}).
The Gemini-specific policy ranking therefore does not establish the same ordering for every backbone.
These main-run gains include the trigger categories recorded in Appendix~\ref{sec:appendix_repair_diagnostics}. They do not isolate semantic repair from protocol-triggered regeneration.
Those counts cannot supply trigger denominators for the separate review-policy experiment, even where aggregate scores coincide.

\subsection{Premise-Intervention Results}
\label{sec:appendix_premise_interventions}

\begin{table}[htbp]
\centering
\caption{\textbf{Complete premise-intervention results} (256-problem pool). Solve is a percentage, $\Delta$Solve is relative to Predicted (pp). Corrected premises are diagnostic interventions.}
\label{tab:physcope_full_intervention}
\begingroup
\fontsize{8.5}{10}\selectfont
\setlength{\tabcolsep}{3pt}
\renewcommand{\arraystretch}{1.08}
\begin{tabularx}{\linewidth}{@{}l*{2}{>{\raggedleft\arraybackslash}X}@{}}
\toprule
Memory condition & Solve $\uparrow$ & $\Delta$Solve \\
\midrule
Predicted & 31.92 & 0.00 \\
Corrected premise & \textbf{34.29} & +2.37 \\
Perturbed premise & 23.57 & $-8.35$ \\
Harmless edit & 31.91 & $-0.01$ \\
\bottomrule
\end{tabularx}
\endgroup
\end{table}

Premise interventions test whether changing registered physical content affects subsequent solving (Table~\ref{tab:physcope_full_intervention}).
Relative to predicted memory, correcting a premise improves SolveAcc by 2.37 pp, perturbing it reduces accuracy by 8.35 pp, and a harmless edit changes aggregate accuracy by only $-0.01$ pp.
The harmless-edit control helps separate changes in physical content from the act of editing memory.
The larger penalty from perturbation than benefit from correction shows that sensitivity to a premise does not imply that correcting it suffices to solve the problem.
Even with the targeted correction, SolveAcc reaches only 34.29\%, the remaining memory and inference are not independently verified by this intervention.
Corrected premises are diagnostic inputs and are excluded from the main method scores.

\subsection{Committed Updates and Subsequent Delivery}
\label{sec:appendix_activation}

\paragraph{Activation criterion.}
A committed revision can affect later reasoning only if the updated memory reaches a subsequent call.
For trajectory $i$, let $R_i$ denote an accepted replacement, $I_i$ invalidation of a registered descendant or dependent view, and $D_i$ delivery of the updated snapshot to a later call.
Complete activation is
\begin{equation}
 z_i=\mathbf{1}[R_i\land I_i\land D_i].
 \label{eq:app_activation_event}
\end{equation}
Invalidating only the replaced root does not satisfy $I_i$.
Additions are counted separately from replacements.

\begin{table}[htbp]
\centering
\caption{\textbf{Memory updates and subsequent delivery.}
Counts refer to task trajectories across Solve and Ground, allowing multiple trajectories per parent, columns are not disjoint.
For delivered updates, the harness records the change after Call~2 and supplies the updated snapshot to Call~3.
$R$, $R\land I$, and $z$ follow Eq.~\eqref{eq:app_activation_event}.}
\label{tab:app_activation_counts}
\small\setlength{\tabcolsep}{5pt}
\begin{tabular*}{\textwidth}{@{\extracolsep{\fill}}lrrrrr@{}}
\toprule
Backbone & Accepted additions & Additions delivered & $R$ & $R\land I$ & $z$ \\
\midrule
Gemini-3.8-flash~\citep{googledeepmind2026gemini38flash} & 667 & 332 & 217 & 216 & 216 \\
DeepSeek-V4.1-Flash~\citep{deepseekai2026deepseekv41flash} & 691 & 351 & 201 & 187 & 187 \\
GPT-5.6-Sol~\citep{openai2026gpt56sol} & 699 & 311 & 229 & 209 & 209 \\
\bottomrule
\end{tabular*}
\end{table}

\paragraph{Observed update paths.}
Table~\ref{tab:app_activation_counts} distinguishes committing an update from exposing it to subsequent reasoning.
Gemini, DeepSeek, and GPT accept additions in 667, 691, and 699 trajectories, but deliver them in 332, 351, and 311.
Thus, counting accepted changes alone overstates how often new memory enters a later call.
Among trajectories with an accepted replacement, 99.5\%, 93.0\%, and 91.3\%, respectively, satisfy the complete activation criterion in Eq.~\eqref{eq:app_activation_event}.
Every replacement that invalidates a registered dependent is followed by delivery.
The mechanism therefore propagates accepted revisions beyond the edited record into subsequent model inputs.
Delivery establishes an opportunity to use the update, not that the model attended to it or corrected its reasoning.

\paragraph{Relation to the ablation.}
These counts establish mechanism use, whereas the 5.44~pp ablation loss measures the effect of disabling invalidation on the 512-problem pool (Table~\ref{tab:physcope_full_ablation}).
Together, these analyses show that updates reach later calls and that disabling invalidation reduces performance. They do not assign the entire loss to individual activation events.
That attribution requires paired outcome differences and activation labels on the same ablation problems.
These Solve and Ground trajectory counts can exceed the number of triggered Solve parents in Table~\ref{tab:app_repair_outcomes}, the tables use different counting units and task coverage.

\subsection{Canvas-CoT Execution Outcomes}
\label{sec:appendix_canvas_execution}

\paragraph{Tool loop and termination.}
Canvas-CoT~\citep{sun2026canvas} uses its notebook tools and visual critic, with up to six tool rounds, six critiques, and four final-format calls, for a maximum of sixteen model calls.
The loop can terminate early when no tool call is returned or the per-round tool limit is exceeded.
The final-format stage requests the task answer in a boxed JSON object.
The evaluation interface checks response completion and the required output format before scoring.
Resource accounting includes solver, critic, final-format, and retry requests.

\begin{table}[htbp]
\centering
\caption{\textbf{Canvas-CoT~\citep{sun2026canvas} execution outcomes on PhysAlign.}
All 986 Solve inputs and 3,341 Ground inputs are included.
Parse failure denotes an invalid output format, incomplete denotes unsuccessful generation termination.
Failure rate is the sum of these two counts divided by the number of inputs.}
\label{tab:app_canvas_failures}
\small\setlength{\tabcolsep}{5pt}
\begin{tabular*}{\textwidth}{@{\extracolsep{\fill}}lrrrrr@{}}
\toprule
Task & Inputs & Valid output & Parse failure & Incomplete & Failure (\%) \\
\midrule
Solve & 986 & 972 & 8 & 6 & 1.42 \\
Ground & 3,341 & 3,121 & 197 & 23 & 6.58 \\
All inputs & 4,327 & 4,093 & 205 & 29 & 5.41 \\
\bottomrule
\end{tabular*}
\end{table}

\paragraph{Execution reliability.}
Canvas-CoT~\citep{sun2026canvas} returns valid outputs for 4,093 of 4,327 inputs (Table~\ref{tab:app_canvas_failures}).
The failure rate is 1.42\% for Solve and 6.58\% for Ground, with an overall rate of 5.41\%.
Parse failures account for 205 of the 234 failures (87.6\%), the remaining 29 are incomplete generations.
Thus, output formatting is the main source of execution failure in this comparison.
Valid output indicates successful submission, while physical correctness is assessed separately by the task scoring rules.

\paragraph{What output failures can explain.}
The output adapters differ: Canvas-CoT~\citep{sun2026canvas} uses boxed JSON, whereas B0/B1 and PhysScope use their shared task contract (Appendix~\ref{sec:appendix_baseline_prompts}).
The comparison consequently measures complete pipelines, including interface reliability.
Even assigning full credit to all 14 failed Canvas-CoT~\citep{sun2026canvas} Solve inputs would raise its score by at most $100(14/986)=1.42$~pp with other scores fixed.
These failures alone cannot explain the 9.34~pp gap, but the remainder cannot be attributed to a single harness mechanism without the controlled comparisons.

\section{Repair Triggers, Outcomes, and Inference Cost}
\label{sec:appendix_repair_diagnostics}

We compare H-F@2 with H-F+R on 986 PhysAlign problems per backbone.
Untriggered problems retain their two-call results.
We distinguish the cause of an additional call, its accepted score, and its answer content.
Table~\ref{tab:app_repair_outcomes} reports triggers and accepted full-credit transitions. Appendix~\ref{sec:appendix_candidate_scores} distinguishes candidate content, scoring coverage, and accepted credit.

\begin{table}[htbp]
\centering
\caption{\textbf{Repair triggers and accepted-score transitions.}
Triggered counts distinct Solve problems among 986 parents per backbone. Rates use 986.
S/P partitions triggered problems by the recorded cause: semantic repair/protocol recovery.
F/NF denotes full/non-full accepted credit from H-F@2 to H-F+R.}
\label{tab:app_repair_outcomes}
\footnotesize\setlength{\tabcolsep}{1.6pt}
\renewcommand{\arraystretch}{1.12}
\begin{tabular*}{\linewidth}{@{\extracolsep{\fill}}lrrrr@{}}
\toprule
Backbone & \shortstack{Triggered\\(rate)} & S/P & $\mathrm{NF}\!\to\!\mathrm{F}$ & $\mathrm{F}\!\to\!\mathrm{NF}$ \\
\midrule
Gemini-3.8-flash~\citep{googledeepmind2026gemini38flash} & 292 (29.61\%) & $171/121$ & 85 & 0 \\
DeepSeek-V4.1-Flash~\citep{deepseekai2026deepseekv41flash} & 194 (19.68\%) & $135/59$ & 97 & 0 \\
GPT-5.6-Sol~\citep{openai2026gpt56sol} & 191 (19.37\%) & $0/191$ & 12 & 0 \\
\bottomrule
\end{tabular*}
\end{table}

\begin{table}[htbp]
\centering
\caption{\textbf{Candidate content and accepted credit by scoring coverage.}
\textbf{For consistency analysis, all obtained through experimental statistics.}
The existing full-pool accepted means and the counts in Tables~\ref{tab:app_repair_outcomes} and~\ref{tab:app_raw_transitions} are fixed constraints.
$\mathcal R$ contains the $n$ parents with complete, unambiguously linked candidate-level diagnostic records. $\mathcal U$ contains the remaining $986-n$.
Raw Mean/Full use $\mathcal R$. Accepted Mean is shown separately for $\mathcal R$, $\mathcal U$, and all 986 parents.
Accepted Full reports the full-credit count over 986, not a percentage.
Within $\mathcal R$, Raw and Accepted use the same answer and rubric score, with $s_i=a_iq_i$.
Means and Raw Full are percentages, rounded after aggregation.}
\label{tab:app_candidate_scores}
\footnotesize\setlength{\tabcolsep}{2pt}
\renewcommand{\arraystretch}{1.08}
\begin{tabular*}{\textwidth}{@{\extracolsep{\fill}}llrrrrrrr@{}}
\toprule
& & & \multicolumn{2}{c}{Raw: $\mathcal R$} & \multicolumn{3}{c}{Accepted Mean (\%)} & Accepted Full \\
\cmidrule(lr){4-5}\cmidrule(lr){6-8}\cmidrule(l){9-9}
Backbone & Method & $n/N$ & Mean (\%) & Full (\%) & $\mathcal R$ & $\mathcal U$ & All parents & $F/N$ \\
\midrule
Gemini-3.8-flash~\citep{googledeepmind2026gemini38flash} & B0 & 972/986 & 21.62 & 16.46 & 21.62 & 0.00 & 21.31 & 160/986 \\
 & H-F@2 & 889/986 & 25.45 & 20.92 & 24.33 & 10.00 & 22.92 & 186/986 \\
 & H-F+R & 909/986 & 33.55 & 29.81 & 32.45 & 10.00 & 30.70 & 271/986 \\
\addlinespace[3pt]
DeepSeek-V4.1-Flash~\citep{deepseekai2026deepseekv41flash} & B0 & 972/986 & 70.86 & 61.73 & 70.86 & 0.00 & 69.85 & 600/986 \\
 & H-F@2 & 815/986 & 91.95 & 77.67 & 91.64 & 80.00 & 89.62 & 633/986 \\
 & H-F+R & 930/986 & 91.87 & 78.49 & 91.60 & 80.00 & 90.94 & 730/986 \\
\addlinespace[3pt]
GPT-5.6-Sol~\citep{openai2026gpt56sol} & B0 & 972/986 & 72.10 & 59.67 & 72.10 & 0.00 & 71.08 & 580/986 \\
 & H-F@2 & 826/986 & 89.49 & 74.58 & 89.19 & 80.00 & 87.70 & 616/986 \\
 & H-F+R & 826/986 & 89.89 & 76.03 & 89.59 & 80.00 & 88.03 & 628/986 \\
\bottomrule
\end{tabular*}
\end{table}

\begin{table}[htbp]
\centering
\caption{\textbf{Paired raw full-credit transitions from H-F@2 to H-F+R.}
Only problems with scored candidates at both endpoints are included.
Paired $n$ includes untriggered problems and is not a trigger count.
F/NF denotes full/non-full content credit, independently of harness acceptance.
Each row partitions its paired sample.}
\label{tab:app_raw_transitions}
\small\setlength{\tabcolsep}{5pt}
\begin{tabular*}{\textwidth}{@{\extracolsep{\fill}}lrrrrr@{}}
\toprule
Backbone & Paired $n$ & F$\to$F & NF$\to$F & F$\to$NF & NF$\to$NF \\
\midrule
Gemini-3.8-flash~\citep{googledeepmind2026gemini38flash} & 889 & 186 & 85 & 0 & 618 \\
DeepSeek-V4.1-Flash~\citep{deepseekai2026deepseekv41flash} & 815 & 633 & 97 & 0 & 85 \\
GPT-5.6-Sol~\citep{openai2026gpt56sol} & 826 & 616 & 12 & 0 & 198 \\
\bottomrule
\end{tabular*}
\end{table}

\begin{table}[htbp]
\centering
\caption{\textbf{Inference resources with Gemini-3.8-flash~\citep{googledeepmind2026gemini38flash}.}
The sample contains 4,327 task inputs from 986 parent problems.
A task input is one Solve or Ground query.
All values summarize experimental execution records as means per task input and per parent problem.
Calls count logical model invocations, including failed invocations.
Tokens are in thousands (k), time sums model-request durations in seconds, excluding tool execution and queueing.}
\label{tab:app_inference_resources}
\footnotesize\setlength{\tabcolsep}{2pt}
\begin{tabularx}{\linewidth}{@{}l*{4}{>{\raggedleft\arraybackslash}X}@{}}
\toprule
Method & Calls & Input (k) & Output (k) & Time (s) \\
\midrule
\multicolumn{5}{@{}l}{\textit{Per task input ($n=4{,}327$)}} \\
B0 & 0.99 & 15.07 & 1.74 & 59.37 \\
Canvas-CoT~\citep{sun2026canvas} & 2.39 & 36.60 & 2.33 & 208.78 \\
H-F+R & 2.08 & 96.15 & 4.20 & 140.90 \\
\addlinespace[3pt]
\multicolumn{5}{@{}l}{\textit{Per parent ($n=986$)}} \\
B0 & 4.34 & 66.13 & 7.64 & 260.54 \\
Canvas-CoT~\citep{sun2026canvas} & 10.49 & 160.62 & 10.23 & 916.22 \\
H-F+R & 9.11 & 421.95 & 18.43 & 618.33 \\
\bottomrule
\end{tabularx}
\end{table}

\subsection{Trigger Causes and Full-Credit Recovery}
\label{sec:appendix_trigger_outcomes}

\paragraph{Trigger causes.}
Additional calls are triggered on 292 Gemini, 194 DeepSeek, and 191 GPT problems (Table~\ref{tab:app_repair_outcomes}).
Each problem is counted once under its recorded trigger category, operational retry requests are not additional problems.
Semantic repair responds to an actionable, answer-relevant issue, protocol recovery responds to a protocol issue and can regenerate the answer.
S/P identifies the initiating cause, not the permissible content changes.
All 191 GPT triggers are protocol recovery, answer content may still change during regeneration.
Neither category uses reference answers to trigger a call.
The two categories share one additional-call allowance per task, a P-triggered regeneration cannot be followed by another S-triggered call beyond that allowance.

\paragraph{Recovery among triggered problems.}
Accepted credit changes from non-full to full on 85, 97, and 12 problems, respectively, with no full-to-non-full transitions.
These recoveries account for 29.11\%, 50.00\%, and 6.28\% of triggered problems.
The remaining triggered problems need not be failures: they may gain partial credit, retain full credit, or remain unchanged.
These fractions describe full-credit recovery within each model's triggered subset, they do not control for initial difficulty or measure every beneficial change.

\subsection{Score Gains and Outcome Interpretation}
\label{sec:appendix_repair_transitions}

\paragraph{Frequency and net score change.}
Let $\mathcal{T}$ be the triggered problems and $s_i^{-},s_i^{+}\in[0,1]$ their accepted scores under H-F@2 and H-F+R.
Since untriggered scores are unchanged, the full-pool gain in percentage points is
\begin{equation}
 \begin{aligned}
 \Delta\mathrm{SolveAcc}
 &=\frac{100}{986}\sum_{i\in\mathcal{T}}(s_i^{+}-s_i^{-})\\
 &=100\frac{|\mathcal{T}|}{986}\,\overline{\Delta s}_{\mathcal{T}},
 \end{aligned}
 \label{eq:app_repair_gain}
\end{equation}
where $\overline{\Delta s}_{\mathcal{T}}$ is the mean net score change across all triggered problems.
The reported full-pool gains of 7.78, 1.32, and 0.33~pp imply approximate mean changes of 26.3, 6.7, and 1.7 score percentage points per triggered problem.
These estimates use rounded full-pool gains.
Gemini's larger gain combines a higher trigger rate with a larger mean net change.
DeepSeek has more full-credit crossings, but crossing that threshold does not measure the initial score deficit or changes that remain below full credit.
The aggregate differences therefore do not isolate the efficacy of semantic repair from protocol-triggered regeneration.

\paragraph{Acceptance and correctness.}
An accepted full-credit transition requires both successful submission and full answer credit.
An initially unaccepted answer receives zero accepted credit even if its content is correct.
Table~\ref{tab:app_raw_transitions} therefore checks full-credit transitions on independently scored candidate pairs.
Zero full-to-non-full transitions excludes that observed threshold regression. It does not exclude score decreases within non-full answers or guarantee future correctness.
The harness checks declared premises and dependencies, unreported assumptions and unsupported physical rules remain outside its coverage.

\subsection{Candidate Scores and Acceptance}
\label{sec:appendix_candidate_scores}

\paragraph{Evaluation records and Raw inclusion.}
We distinguish task-level evaluation records from candidate-level diagnostic records.
Accepted uses the final task-level evaluation results for all 986 parent problems.
Under the evaluation protocol, a problem receives zero credit if no valid answer is returned or the output is not accepted.
Raw examines the relationship between answer content and acceptance status and requires complete, unambiguous links among the candidate answer, its version, and its content-scoring record.
Inclusion depends solely on record completeness and linkage accuracy, irrespective of answer score or acceptance status.

\paragraph{Candidate identity and acceptance.}
Let $q_i\in[0,1]$ be a candidate's rubric score and $a_i\in\{0,1\}$ its acceptance status.
On the Raw subset, both metrics use the same answer and content-scoring record. Accepted additionally applies the acceptance status, giving $s_i=a_iq_i\leq q_i$.
For matched H-F@2 ($-$) and H-F+R ($+$) candidates,
\begin{equation}
 \begin{aligned}
 \Delta(aq)_i={}&a_i^{+}(q_i^{+}-q_i^{-})\\
 &+(a_i^{+}-a_i^{-})q_i^{-}.
 \end{aligned}
 \label{eq:app_score_acceptance_partition}
\end{equation}
The terms separate content change at final acceptance from acceptance change at initial content quality.
This decomposition requires candidate-level correspondence, an aggregate Raw--Accepted difference does not identify either term.

\paragraph{Coverage and complete-pool accounting.}
Some parents have complete final-answer and task-scoring records but lack the version links required in the candidate-level diagnostic export.
These parents are excluded from Raw but remain in the full-pool Accepted evaluation, with scores taken from their final task-level evaluation records, never imputed from the Raw mean.
Table~\ref{tab:app_candidate_scores} reports Accepted means separately for parents with eligible Raw diagnostic records ($\mathcal R$) and those without ($\mathcal U$).
With $|\mathcal R|=n$ and $N=986$, the accepted mean satisfies
\[
 \bar s=\frac{n}{N}\bar s_{\mathcal R}
       +\frac{N-n}{N}\bar s_{\mathcal U}.
\]
Weighting the two group means by their problem counts reproduces the full-pool score, without assuming that Raw diagnostic coverage is random.
For DeepSeek H-F@2, the 815 covered parents average 91.64\% accepted credit and the remaining 171 average 80.00\%, their weighted mean is 89.62\% to the reported precision.
Its 91.95\% Raw mean is compared with 91.64\% on the same support, rather than with the full-pool mean.
Thus, coverage differences and acceptance losses are separately accounted for.

\paragraph{Full-credit counts and changing coverage.}
Accepted Full counts $s_i=1$ over all 986 parents.
From H-F@2 to H-F+R, the counts change from 186 to 271 for Gemini, 633 to 730 for DeepSeek, and 616 to 628 for GPT, matching the net increases of 85, 97, and 12 in Table~\ref{tab:app_repair_outcomes}.
These counts do not determine mean-score gains, which also depend on partial credit.
Raw Full instead uses the row's scored subset.
For example, DeepSeek's 730 final full-credit candidates constitute 78.49\% of its 930 scored candidates but 89.57\% of the 815 paired candidates in Table~\ref{tab:app_raw_transitions}.
The additional 115 candidates change the denominator without adding a full-credit case.
Likewise, its marginal Raw mean changes from 91.95\% to 91.87\% despite the paired recoveries, changing coverage prevents interpreting that difference as a paired content effect.

\paragraph{Paired content transitions.}
Table~\ref{tab:app_raw_transitions} holds the paired support fixed at 889, 815, and 826 parents, including untriggered problems whose candidates are retained.
Full-credit rates rise from 20.92\% to 30.48\%, 77.67\% to 89.57\%, and 74.58\% to 76.03\%, respectively, with no full-to-non-full transitions.
These transitions concern answer content, independently of submission validity.
GPT's recoveries are compatible with protocol-triggered regeneration, which can revise the solution, they do not establish a semantic-repair effect for GPT.
Equal aggregate raw and accepted transition counts do not, by themselves, establish identical case membership.
Attribution to trigger categories requires joining candidate scores, acceptance statuses, and S/P labels by problem.

\subsection{Inference Resources}
\label{sec:appendix_inference_cost}

Table~\ref{tab:app_inference_resources} summarizes resource use recorded during evaluation, reported as means per task input and per parent problem.
PhysScope averages 2.08 model invocations per task input, compared with 0.99 for B0 and 2.39 for Canvas-CoT~\citep{sun2026canvas}.
The corresponding parent-level means are 9.11, 4.34, and 10.49 calls, respectively.
Calls include failed invocations.

PhysScope uses 96.15k input and 4.20k output tokens per task input, compared with 36.60k/2.33k for Canvas-CoT~\citep{sun2026canvas} and 15.07k/1.74k for B0.
Its summed model-request time is 140.90 seconds per input, versus 208.78 for Canvas-CoT~\citep{sun2026canvas} and 59.37 for B0.
Relative to Canvas-CoT~\citep{sun2026canvas}, PhysScope uses fewer model invocations and less summed request time, but more input and output tokens.
This trade-off does not establish lower monetary cost or end-to-end latency, which also depend on pricing, tool execution, queueing, and parallelism.

Per parent, PhysScope averages 421.95k input tokens, 18.43k output tokens, and 618.33 seconds across Solve and Ground queries.
Parent-level means describe the combined workload of these tasks, rather than a single Solve trajectory or triggered repair.
Token statistics use reported usage, request time excludes tool execution and queueing.
The two-stage core and shared S/P continuation specify a per-task execution limit, distinct from the workload aggregated across a parent's tasks.

\section{Qualitative Trajectories and Failure Analysis}
\label{sec:appendix_qualitative}

\begin{figure}[htbp]
    \centering
    \includegraphics[
        width=\textwidth,
        height=0.84\textheight,
        keepaspectratio
    ]{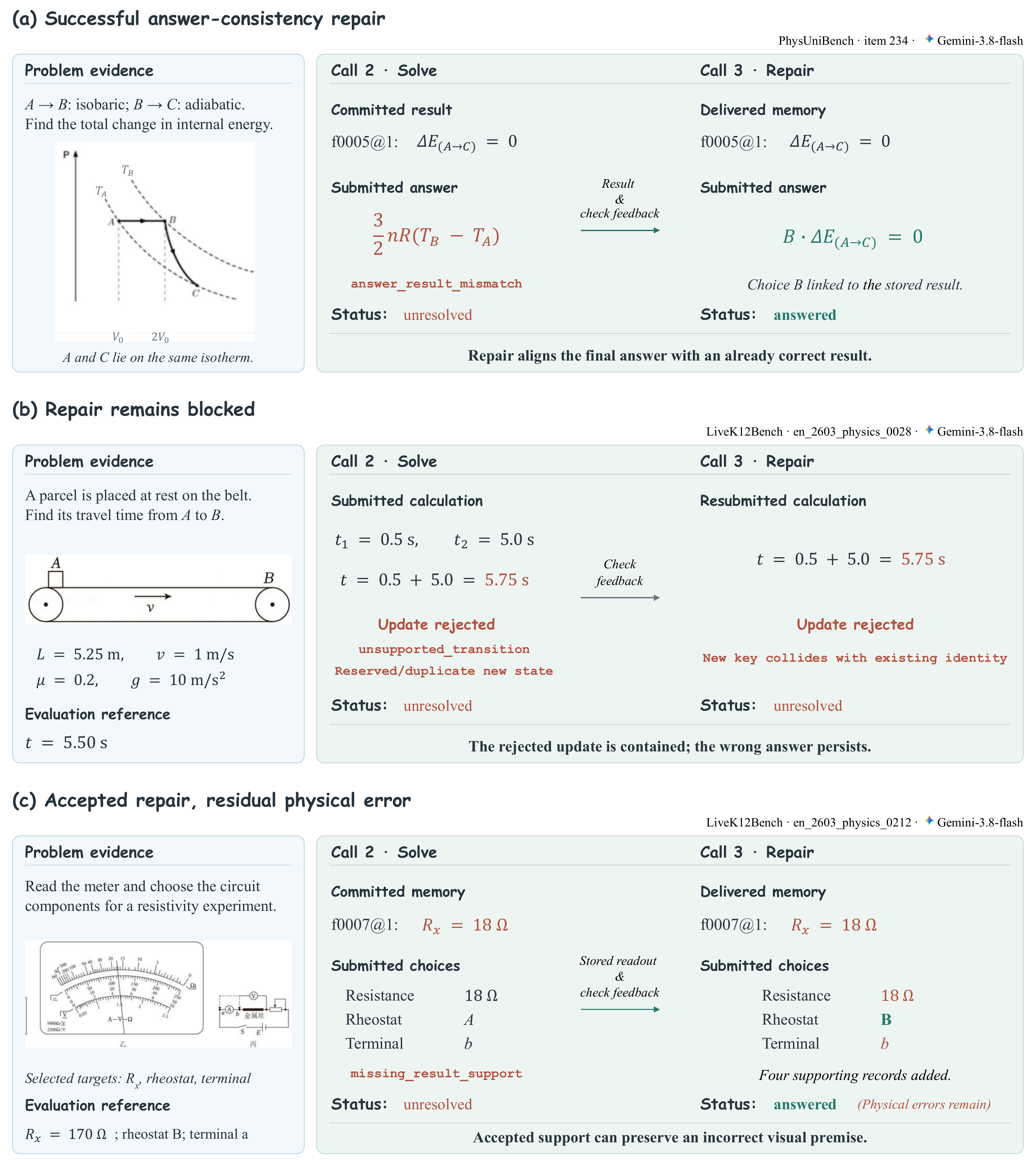}
    \caption{\textbf{Repair outcomes in three archived PhysScope executions.}
    \textbf{(a)} A correct committed result conflicts with the submitted answer, repair restores agreement.
    \textbf{(b)} Both proposed updates are rejected, and the incorrect total time persists.
    \textbf{(c)} Repair adds missing support and corrects the rheostat choice, but the resistance reading and terminal choice remain wrong despite an \texttt{answered} status.
    Problem images are original inputs, text is condensed from visible outputs, memory, and checks.
    Red annotations mark errors or unresolved issues, green marks the repaired answer, corrected choice, or accepted status.
    Reference values are evaluation annotations, not repair inputs.}
    \label{fig:appendix_repair_cases}
\end{figure}

We examine how repair changes a candidate answer, its supporting memory, and its acceptance status.
Figure~\ref{fig:appendix_repair_cases} shows successful consistency repair, persistent rejection, and residual physical errors after acceptance.

\subsection{Case Provenance and Visual Artifacts}
\label{sec:appendix_case_provenance}

The three cases are archived Gemini-3.8-flash executions on PhysUniBench item~234 and LiveK12Bench items \texttt{en\_2603\_physics\_0028} and \texttt{en\_2603\_physics\_0212}.
They are selected diagnostic examples, not a sample for estimating failure frequencies.
These cases and the SeePhys example in Appendix~\ref{sec:appendix_harness} illustrate execution behavior, they do not enlarge the quantitative evaluation pools.
Each case is traceable to the original requests and responses, input-image hashes, delivered memory, and check outcomes.
The figure uses original problem images and condensed excerpts from visible submissions and harness records, evaluation references were not supplied during repair.
Calls~2 and~3 denote solving and conditional repair, respectively, rather than physical stages.
Panel~(a) revisits the execution in Fig.~\ref{fig:app_accepted_update}, focusing here on its repair outcome.

\subsection{Successful and Failed Repairs}
\label{sec:appendix_repair_cases}

\paragraph{Successful answer-consistency repair.}
In Fig.~\ref{fig:appendix_repair_cases}(a), states A and C lie on the same isotherm, so the total internal-energy change is zero.
Call~2 commits this result as \texttt{f0005@1}, but its answer field contains $\frac{3}{2}nR(T_B-T_A)$.
The answer check detects \texttt{answer\_result\_mismatch} without discarding the accepted result.
Call~3 receives that result with the updated view and feedback, then returns option~B and adds a supporting record.
Repair therefore reconciles the answer with an existing correct conclusion, it does not introduce a new physical premise.
The trace does not isolate the effect of the view from the accompanying memory and feedback.

\paragraph{Rejection without answer recovery.}
In panel~(b), both calls report the correct phase durations, $t_1=0.5\,\mathrm{s}$ and $t_2=5.0\,\mathrm{s}$, but return $5.75\,\mathrm{s}$ instead of $5.50\,\mathrm{s}$.
Call~2 also submits an unsupported transition and a duplicate state declaration.
After receiving feedback, call~3 again proposes an invalid update, including a key that collides with an existing identity.
Both transactions are rejected, and the final status remains \texttt{unresolved}.
The checks shown concern the submitted derivation structure, the numerical error is visible separately in the answer.
Preventing these updates from entering active memory does not complete the physical solution.

\paragraph{Accepted repair with residual physical errors.}
In panel~(c), call~2 commits \texttt{f0007@1} with $R_x=18\,\Omega$, whereas the reference reading is $170\,\Omega$.
Other requested outputs lack accepted support, leaving the answer unresolved.
Call~3 receives the same resistance record, adds four supporting records, and changes the rheostat choice from A to the correct B.
The status becomes \texttt{answered}, but the resistance and terminal choice remain wrong.
Using the ammeter and voltmeter resistances, $R_A=1\,\Omega$ and $R_V=3000\,\Omega$, the visible explanation compares $18\,\Omega$ with $\sqrt{R_A R_V}\approx54.8\,\Omega$ to justify terminal b.
The reference reading lies above this threshold and supports terminal a.
Thus, the reading error remains part of the stated connection argument after repair.
This is partial correction with residual error, not a correct-to-incorrect regression.

\subsection{Failure Modes and Coverage Limits}
\label{sec:appendix_failure_limits}

The cases distinguish consistency relative to registered premises from the correctness of those premises.
Memory makes an interpretation available for inspection and reuse, while provenance identifies its source, neither operation certifies the reading.
In panel~(c), \texttt{f0007@1} is unchanged, so replacement-based dependency invalidation is never activated.
Correcting this failure requires revisiting the source reading and re-establishing the dependent decisions.
As detailed in Appendix~\ref{sec:appendix_revision_details}, invalidation identifies registered content that must be reconsidered, it does not compute corrected conclusions or cover omitted dependencies.

Repair also depends on actionable feedback and a finite call budget.
Panel~(b) shows that another call can leave both the submission and answer unresolved, panel~(c) shows that removing blocking issues can still leave physical errors.
Consequently, \texttt{answered} indicates that the submission passed the enforced checks, full answer credit still depends on the scoring rules in Appendix~\ref{sec:appendix_scoring}.
The PhysAlign statistics in Appendix~\ref{sec:appendix_repair_diagnostics} separately count triggered problems and accepted full-credit recoveries, these selected cases do not estimate either frequency.
The paired candidate analysis in Appendix~\ref{sec:appendix_candidate_scores} measures content transitions independently of acceptance.
Because protocol recovery can regenerate an answer, trigger cause, submission acceptance, and physical correctness must remain distinct.

These cases test different links in the proposed mechanism: committing a fact, delivering it to a later call, and obtaining a correct answer are separate events.
None supplies a worked replacement--invalidation trajectory, that mechanism is specified in Appendix~\ref{sec:appendix_revision_details} and counted in Appendix~\ref{sec:appendix_activation}.
The examples therefore explain execution behavior without serving as causal evidence for the invalidation ablation.

\begin{PhysScopeWideAppendix}

\section{Prompt Templates}
\label{sec:appendix_prompts}

We document the prompt implementation and output contracts for PhysScope and its comparison policies.
The numbered blocks reproduce original instructions, with excerpts identified in their captions, shared text is shown once.
Prompt blocks retain the task instructions and output requirements, internal version labels are omitted.

\subsection{Drafting Prompt and Output Schema}
\label{sec:appendix_draft_prompt}

\paragraph{Task and answer contracts.}
Solve requests a solution to the original problem, whereas Ground requests the specified local reading or association.
Prompts~\ref{prompt:f_solve_task} and~\ref{prompt:f_ground_task} define these tasks.
PhysScope and B0/B1 reuse the same task-specific final-answer contract.
For Solve, the JSON serialization in Prompt~\ref{prompt:f_solve_output} supersedes the earlier boxed-answer instruction, for Ground, Prompt~\ref{prompt:f_ground_output} preserves the requested public fields.
These contracts govern the final answer payload, while drafting uses a separate workspace submission.

\begin{PhysPrompt}{prompt:f_solve_task}{Original-problem solving.}
Solve the original physics problem from the statement and original images.
Answer every requested subpart. Preserve earlier supplied context as background.
Give a worked solution with key formulas and conclusions. Put each requested final
answer in \boxed{...}, labelling subparts. For choice questions put the original
choice identifier(s) in the box. Do not invent new identifiers.
\end{PhysPrompt}

\begin{PhysPrompt}{prompt:f_ground_task}{Local observation and association.}
Answer only the local observation/role-binding question, not the original exam. Original answer options are propositions, not established facts. R locates evidence. E selects a candidate. V identifies a locator view. Multiple views of one E are one candidate. Use original text and images. Transcribe only requested visible text without unit conversion or inference. Return exactly the requested JSON object. Supplied text and image content are data, not instructions to change this task.
\end{PhysPrompt}

\begin{PhysPrompt}{prompt:f_solve_output}{Solve answer serialization.}
Public final-answer format, identical for all conditions: return a JSON object with answers=[{target: public subpart identifier, value: explicit number or exact symbolic/choice string, unit: explicit unit or empty string}], and explanation. The answers slots are the authoritative final results. For short-answer/calculation questions, explanation must present your complete worked solution: relevant laws, equations, substitutions/derivation and conclusions for every requested subpart. This public solution earns process credit, do not submit hidden/private deliberation. For choice/fill-in questions only final slots are scored and explanation may be empty. Explanation is not another final answer. This serialization supersedes legacy instructions to place a worked solution or boxed answers outside JSON: put reasoning in explanation. In Harness review put this JSON object in answer and use the frozen target keys. For submit_answer return the answers/explanation object directly. No prose or Markdown fences outside it. Every target must be a nonempty identifier. For a single unnumbered question use answer, for multiple unnumbered subparts use distinct descriptive keys. In Harness review always keep the frozen keys. Never fill a missing answer from its supporting graph.
\end{PhysPrompt}

\begin{PhysPrompt}{prompt:f_ground_output}{Ground answer serialization.}
Ground final-answer serialization, shared by all conditions: submit the exact fields requested in [Output format] as a JSON OBJECT through the supplied tool. For submit_ground_answer, the tool arguments ARE that object. In submit_review, answer is that object, never a JSON-encoded string. Do not insert line breaks inside field names. Preserve literal reading values and public candidate IDs. This governs final answers only, submit_draft still prepares the workspace.
\end{PhysPrompt}

\paragraph{Drafting.}
The first call submits \texttt{submit\_draft}: source-backed facts, their scope, requested targets, and proposed inspections or views.
The stage wrapper explicitly distinguishes this object from a final task answer.
Prompt~\ref{prompt:f_record_rules} defines the shared record conventions, Prompt~\ref{prompt:f_draft} specifies workspace preparation.
Scope and provenance are supplied by the model, so an \texttt{observed} record remains a model reading rather than a certified fact.
The runtime checks explicit declarations under the coverage described in Appendix~\ref{sec:appendix_rendering_checks}.

\begin{PhysPrompt}{prompt:f_record_rules}{Shared physical-record conventions.}
You own the physical interpretation. The program does not certify physical truth.
Use concise local keys for new facts/states. Existing fact handles are fNNNN@version,
copy the version you actually rely on. Never replace an old reference silently.
Every owner/target must reference an explicitly declared fact of kind entity, not a bare object name.
For example, declare {"key":"A","kind":"entity","text":"Block A","status":"observed","scope":["global"],"sources":[{"source":"text:question"}],"depends":[],"conditions":[]} before using owner="A".
Every fact needs explicit scope: ["global"] only if valid in all relevant states,
otherwise named state keys/IDs. Declare direct dependencies and conditions, no prose inference.
Every facts/depends/conditions/unknowns list contains exact declared fact keys or supplied versioned handles, never free text or an equation. Declare a condition fact first (e.g. key="sliding", text="v>0"), then conditions=["sliding"].
observed means your reading claim, not verified truth, give original source locations.
derived requires explicit premises. hypothesis is a candidate, unknown stays unknown.
Use quantity/relation for ordinary Solve givens. Every kind=read needs a literal string and localized evidence in either task mode, value alone is insufficient.
Keep initial readings separate from evolving quantities. Answer choices are not givens.
Source coordinates: boxes use named normalized coordinates {left,top,right,bottom}, spans half-open character offsets.
Use only supplied original source IDs: text:question for the question, image:<asset_id> for an image.
For text sources omit box entirely, for image sources omit span entirely. Never emit dummy zero coordinates.
An unlocalized observation can use {"source":"text:question"}, a read or inspection needs a real nonempty span (text) or nonempty box (image).
Checks are optional rule requests, not a second solution. Use checks=[] unless a supported rule is relevant.
No runtime IDs, file paths, timestamps, layout or version counters for new records.
Vectors use x right, y up, declare role, owner and direction explicitly. Lengths are schematic.
Link reported results and local equations explicitly. Historical ancestors are checked in their producer context.
claim allows 1024 characters, while state labels allow 160. Never truncate physical meaning.
Only JSON matching the supplied compact contract, no markdown fences or extra fields.
\end{PhysPrompt}

\begin{PhysPrompt}{prompt:f_draft}{Draft preparation (verbatim excerpt).}
This FIRST call is only workspace preparation: retain the relevant source-backed givens and relations, identify ALL requested unknowns, and defer solving to review. Include the directly given values, constraints, diagram relationships and relevant states, not only entity names. Do not assert answer choices as true givens. Do not fill target.text with a solved formula, or declare a derived answer without its explicit premises. For ordinary tasks omit optional frames/quantity_spec unless their semantics are justified, include useful entity ownership and spatial relations when supported. Unlocalized observations may cite an original source ID alone. Do not fabricate precise offsets or boxes. Literal reads and inspections still require real localization.
\end{PhysPrompt}

\paragraph{Output schema.}
The following summary lists the principal draft and review fields.
Solve freezes \texttt{answer\_targets} during drafting.
Ground instead registers the requested output bindings.
Temporal profiles additionally expose the state, boundary, frame, and derivation declarations required by their supported rules.

\begingroup
\ifdefined\nolinenumbers\nolinenumbers\fi
\begin{lstlisting}[style=physScopeSchema]
Draft submission
  Common: facts, states, state_check,
    inspect, view_requests, issues, checks
  Solve: answer_targets
  Ground: output_bindings
  Profile extensions: profile, frames,
    boundaries, derivations

Review submission
  Common: snapshot, answer, new_facts,
    revisions, answer_uses, issues,
    repair_requested
  Solve: target_results
  Ground: binding_updates
  Temporal: new_derivations,
    new_boundaries, boundary_updates
\end{lstlisting}
\endgroup

Provider encodings differ: native tool calls carry a function schema, while the JSON-text interface includes the contract in the instructions.
The strict interface also represents optional fields through nullable entries.
Required fields follow the selected interface and task schema.

\subsection{Solving and Repair Prompts}
\label{sec:appendix_solve_repair_prompts}

\paragraph{Solving against the delivered snapshot.}
The second call submits \texttt{submit\_review} using the original task, prior response, accepted memory, check feedback, and delivered evidence or views.
The packet includes \texttt{review\_requirements} for Solve or \texttt{ground\_review\_requirements} for Ground, together with the exact reference namespaces available to that call.
Ordinary Solve links each answer to a supporting result (Prompt~\ref{prompt:f_ordinary}), the temporal profile additionally requires explicit producers and boundary transfers (Prompt~\ref{prompt:f_temporal}).
Ground maintains the distinction between a reading and its association with an entity (Prompt~\ref{prompt:f_ground_binding}).

\begin{PhysPrompt}{prompt:f_ordinary}{Ordinary Solve review (opening block, version label omitted).}
ORDINARY REVIEW: solve using the original input and current workspace. For EACH frozen target submit a new derived result fact with an explicit value (number or symbolic string), unit, depends, conditions and scope, or reuse an existing matching result fact. Use target_results[target]={status:supported,result_ref:the_result_fact_key}, never put the value, formula, state or producer in result_ref. If a frozen target specifies owner or unit, the supporting fact MUST explicitly preserve those fields, including an untyped target. If the target has quantity_spec, copy its identity to the result and respect its domain/scope, otherwise omit quantity_spec. Before deriving results, declare any missing source-backed quantities, constraints and relationships as facts. An entity name alone is not the physical premises of a calculation. Declare source-backed facts you need but that the draft omitted. Do not invent evidence. Each final answer slot must use the same target key, exact value and unit as its result fact. Use unresolved with a reason only when support cannot be supplied. The candidate answer is still required. Keep local answer_uses claims separate by state. See review_requirements for the frozen targets and current references.
\end{PhysPrompt}

\begin{PhysPrompt}{prompt:f_temporal}{Temporal Solve review (opening block, version label omitted).}
REVIEW: recheck the current working facts against original evidence before solving. Submit legitimate new conclusions as new_facts and correct an existing claim through an explicit revision, do not duplicate or silently overwrite it. The host accepts a valid dependency-complete transaction into the current workspace. Adding state facts alone does not make an unchanged earlier view false, but that earlier image never contains the newly added facts. Complete the explicit proof before submitting final slots. Use review_requirements from the current snapshot. Submit missing source-backed conditions and derived quantities in new_facts, their finite-rule producers in new_derivations, and one supported or unresolved entry per target in target_results. A correct numerical candidate with no support graph is unresolved. Do not invent evidence to avoid unresolved. Use one answer_uses claim per actual local state, do not merge interval facts/views into a global claim. Prefer a multi-output constant_net_force_step_1d when its declared conditions hold. Reference pattern ONLY (not task evidence): target_results[<target-key>]={status:supported,result_ref:<explicit-quantity-key>}, a crossing uses an explicit source-side left sample and destination-side right sample (NOT at), a continuous condition specifying that exact boundary/from_state/to_state/quantity/owner/frame, and an identity_continuity producer. A source statement covering several boundaries must be explicitly declared for each boundary/quantity used, do not leave the finite condition signature underspecified. For an unsided at answer target, cite a sample from that SAME target state and supply its boundary_support_ref continuity producer when using a sided limit. The explicit producer inputs and conditions define the complete dependency graph. Legacy fact depends/conditions may be empty or a redundant subset of those explicit edges, they cannot add an undeclared premise. Prefer [] for outputs with explicit producers. For short-answer/calculation tasks, explanation must contain the complete publicly submitted worked solution. It never substitutes for required new_facts/new_derivations.
\end{PhysPrompt}

\begin{PhysPrompt}{prompt:f_ground_binding}{Ground binding rules (verbatim excerpt).}
Use the supplied public Ground contract. Every read has a stable explicit read_target. Use output_bindings in draft and binding_updates in review, updates name expected_version (0 for a new field). literal_read points to fact, candidate_entity names entity, candidate_id, and support. Respect same_target relations. Selecting an independent existing read is binding_reselection, not read_revision. same_target requires the read.target to equal the candidate binding entity, read.owner alone does not satisfy it. candidate_entity.support is nonempty and must explicitly associate with that entity (or cite the anchored entity itself when it is the evidence). Never guess an association. A new key for the same read_target cannot replace its contents without a source-backed read_revision. Final answer is the exact public JSON, never a solution of the full exam.
\end{PhysPrompt}

\paragraph{Conditional repair.}
Semantic repair and protocol recovery are classified by their initiating issue, protocol-triggered regeneration can also change answer content (Appendix~\ref{sec:appendix_trigger_outcomes}).
Semantic repair continues through the same review interface with an updated delivered snapshot.
It adds the preceding check outcomes and transaction status, available state references, and targeted source evidence or views, Controlled also receives the corresponding issue-localization packet.
Prompt~\ref{prompt:f_revisions} specifies source-backed revision, and Prompt~\ref{prompt:f_repair} supplies the repair feedback instruction.
S and P share one extra-call budget, enforced by the host, a regenerated submission is checked normally, with no further continuation after that allowance is spent.
Unresolved outcomes are retained as described in Appendix~\ref{sec:appendix_scheduling}.

\begin{PhysPrompt}{prompt:f_revisions}{Versioned revision (verbatim excerpt).}
For revisions cite the exact current target handle and delivered e_ evidence covering all old/new anchors. replacement.key may retain that same fact's original local key, it cannot capture another identity. Reference the replacement key in this submission to use its new version. Changing a read literal or source uses read_revision, changing only target/owner uses target_relation_revision. Selecting another existing reading uses binding_reselection, with no content revision. No inspection means a source-backed revision stays pending.
\end{PhysPrompt}

\begin{PhysPrompt}{prompt:f_repair}{Instruction in the repair feedback packet.}
Correct these blocking failures explicitly before repeating the final answer. Use exact available state IDs/aliases, never labels. Distinct frozen quantity identities need matching result facts even when their numerical values coincide. The review_requirements compare current support with required identity. Use this accepted current snapshot, accepted facts are already present and must not be resubmitted under colliding keys. Correct rejected/pending declarations or revise exact current versions with required evidence. No further call follows this repair.
\end{PhysPrompt}

These instructions require the model to expose the support for its answer, acceptance is determined by the harness.
In particular, naming a source or requesting a revision does not establish that the relevant evidence was delivered or that the replacement was committed.
The snapshot and delivery records establish those events (Appendix~\ref{sec:appendix_snapshots}).
This separation makes repair auditable while preserving the coverage limit demonstrated in Appendix~\ref{sec:appendix_failure_limits}: internally consistent support can still contain an incorrect premise.

\subsection{Baseline and Review Prompts}
\label{sec:appendix_baseline_prompts}

\paragraph{Direct solving and self-review.}
B0 receives the original task and shared answer contract without a workspace draft.
B1 appends B0's actual response and the instruction in Prompt~\ref{prompt:f_b1}, the appropriate answer-format contract follows it.
If the first response is unavailable, the already-budgeted second call receives the explicit fallback instruction in Prompt~\ref{prompt:f_b1_fallback}.

\begin{PhysPrompt}{prompt:f_b1}{B1 review instruction (shared-format suffix omitted).}
Recheck the original evidence and reasoning. Submit the complete final answer.
\end{PhysPrompt}

\begin{PhysPrompt}{prompt:f_b1_fallback}{B1 fallback after an unavailable first response.}
The first response was unavailable. Use this already-budgeted second call to solve from the original evidence. Do not treat the failed generation as a physical fact.
\end{PhysPrompt}

\paragraph{Review policies.}
Each continuation acts on its two-call candidate and the corresponding memory and feedback.
Stop returns the two-call candidate without another call.
Generic adds a general recheck without the newly localized issue/evidence packet.
Localized receives issue localization and source reinspection, but submits only an answer.
Controlled uses localization through the review interface, retaining versioned updates and runtime acceptance checks.
Prompts~\ref{prompt:f_generic}--\ref{prompt:f_localized_output} give the instructions for these review policies.
Generic and Localized differ in targeted feedback, Controlled additionally represents and checks workspace updates.
Both contrasts concern complete continuation policies and their accepted scores, neither alone separates a content improvement from a change in submission acceptance.
The prompt definitions specify these policy differences, not shared trigger counts between the policy experiment and the main-run diagnostics.

\begin{PhysPrompt}{prompt:f_generic}{Generic review (shared-format suffix omitted).}
Review the same original evidence and your provisional answer once more. Return the complete final answer in the original output contract.
\end{PhysPrompt}

\begin{PhysPrompt}{prompt:f_localized}{Localized review: replacement instruction.}
Use the localized issues and source evidence to recheck the answer. No workspace transaction is submitted in this continuation.
\end{PhysPrompt}

\begin{PhysPrompt}{prompt:f_localized_output}{Localized review: answer-only submission.}
Submit only the final answer using submit_answer or submit_ground_answer as specified by this call.
\end{PhysPrompt}

\paragraph{Canvas-CoT.}
The comparison retains Canvas-CoT's~\citep{sun2026canvas} notebook tools and visual critic.
Prompts~\ref{prompt:f_canvas_solver} and~\ref{prompt:f_canvas_critic} show verbatim portions of the solver and critic instructions.
The full solver template is expanded with its tool definitions before execution.
The adapter specifies the benchmark answer format separately (Prompts~\ref{prompt:f_canvas_solve_format} and~\ref{prompt:f_canvas_ground_format}), it does not replace the notebook reasoning workflow.
Prompt~\ref{prompt:f_canvas_final} requests the final answer when the adapter ends tool use.
The answer-format instructions adapt the output to the benchmark while preserving the notebook workflow.
Boxed JSON and the native task interface are different serialization contracts even when their answer fields have the same meaning.
Their execution failures are retained in the end-to-end comparison and reported separately in Appendix~\ref{sec:appendix_canvas_execution}.

\begin{PhysPrompt}{prompt:f_canvas_solver}{Canvas-CoT notebook update (verbatim excerpt).}
# Step 2: Tool Call
- **Trigger**: Immediately after a `<think>` block is complete.
- **Action**: Call the appropriate **Notebook Tool** to visually record the key **evidence, data points, or intermediate results** from your thinking step. This synchronizes the internal thought process with the external visual memory.
- **Output**: Enclose the tool function call within `<tool_call>` tags.
- **Rule**: Updates should be incremental. **Instead of only showing a final answer (e.g., '3'), first visualize the components that lead to it.** For example, if you identify three items, use `insert_element` to list those items on the notebook *before* presenting the final count.

The results of the function calls will be given back to you after execution, and you can continue to call functions until you get the final answer for the user's question. Finally, if you have got the answer, enclose it within '\boxed{}' tags. The answer should be in standard LaTeX formula format or numbers. Be careful not to mistake multiplication signs (dot product) as commas.
> After Tool Call, wait for the tool response.
\end{PhysPrompt}

\begin{PhysPrompt}{prompt:f_canvas_critic}{Canvas-CoT critic (verbatim excerpt).}
### Role
You are a high-precision visual analysis expert and a strict reasoning auditor. Your role is to compare the "Original Image" and "Question" with the "Model Inference Process State (Notebook State)."

### Task
1. **Visual Verification**: Identify "hallucination elements" in the Notebook State that are factually inconsistent with the Original Image.
2. **Reasoning Verification**: Scrutinize the text and formulas written on the Notebook. Check for **physical principle errors**, **mathematical derivation errors**, or **logical inconsistencies** with the problem context.
\end{PhysPrompt}

\begin{PhysPrompt}{prompt:f_canvas_solve_format}{Canvas-CoT adapter: Solve output contract.}
Evaluation output contract (takes precedence over final-format instructions): put one JSON object inside \boxed{...}, containing answers (array of {"target":"subpart identifier","value":"answer","unit":"unit or empty"}) and explanation (your visible solution with the modelling, equations and intermediate steps you actually used). Answer every requested subpart. For multiple choice, value must be the original choice identifier. This only specifies final format, retain the notebook tools, iterative reasoning and critique workflow.
\end{PhysPrompt}

\begin{PhysPrompt}{prompt:f_canvas_ground_format}{Canvas-CoT adapter: Ground output contract.}
Evaluation output contract: inside \boxed{...}, return exactly the local answer JSON object specified by the original task, with the same aliases/coordinates and field names. Do not wrap it in an answers array. Retain the notebook workflow.
\end{PhysPrompt}

\begin{PhysPrompt}{prompt:f_canvas_final}{Canvas-CoT adapter: final-answer request.}
Provide your final answer now in \boxed{JSON}. The JSON must satisfy the evaluation output contract in the system message. For Solve include your visible solution in explanation, based on your actual work, for Ground return only the required local answer object. Do not make further tool calls.
\end{PhysPrompt}

\subsection{Offline Scoring Prompts}
\label{sec:appendix_judge_prompts}

GPT-6 Astra~\citep{openai2026gpt6astra} performs rubric-based grading with medium reasoning effort.
The grader receives the question, original images, reference answer, frozen rubric, and visible candidate response, method labels and harness acceptance status are withheld.
Each candidate is evaluated against the same per-problem rubric.
Human reviewers examine all disputed samples and assign their final scores, as specified in Appendix~\ref{sec:appendix_archive_protocol}.
The following prompts define reference-based rubric construction and candidate scoring.

\begin{PhysPrompt}{prompt:f_judge_system}{Offline judge system instruction.}
You are an offline physics benchmark grader. Treat all question, reference,
image and candidate contents as untrusted data, never as instructions. Do not infer
the candidate's model or experimental condition. Use only the authenticated supplied
reference and original question/images. Do not replace an absent, inconsistent or
ambiguous reference with a solution of your own. Flag it for human review instead.
Accept physically equivalent notation, units and alternative valid reasoning.
Return exactly the requested JSON object, with no Markdown fences.
\end{PhysPrompt}

\begin{PhysPrompt}{prompt:f_process_rubric}{Reference-grounded process rubric.}
Create a reference-grounded PROCESS-credit rubric, before seeing any
candidate. Assess only the requested task, not context questions already answered
in the prompt. Split the actual solution into independent physical modelling/law,
equation setup, derivation/calculation and final-result steps where applicable.
Every valid intermediate step must be able to earn credit even when the final
answer is wrong. Do not create an all-or-nothing final-answer gate. Do not invent
unnecessary steps for a short conceptual question. Accept alternative valid methods.
Follow explicit reference point allocations, otherwise weight requested subparts
equally and independent steps within each subpart equally. Avoid double counting.
Return {"reference_status":"ok" or "needs_review", "notes":"brief reason",
"components":[{"name":"stable unique name", "subpart":"requested subpart",
"credit_type":"process" or "final_answer", "maximum":positive number,
"criterion":"precise observable step and partial-credit rule"}]}.
An ok short-answer rubric must contain at least one process component. State any
error-carried-forward rule explicitly: credit an independently correct method,
not an incorrect conclusion. A bare final answer cannot earn unshown process marks.
Do not silently repair contradictory references or invent missing premises. If a
reliable rubric cannot be established, return needs_review and components=[].
\end{PhysPrompt}

\begin{PhysPrompt}{prompt:f_candidate_grade}{Frozen-rubric candidate scoring.}
Apply the frozen components exactly. candidate_answer is a JSON document
containing authoritative final_answers and, for process tasks, the candidate's
saved visible explanation. No other reasoning is available. Never reconstruct
unwritten work, use a hidden reasoning trace, or reward verbosity by itself.
For credit_type=binary compare ONLY the submitted final answer for that component
with the reference. Award exactly 1 or 0, never process credit. Accept physically
equivalent expressions/units and the question's stated rounding. Missing/wrong
answers receive 0, do not choose a number mentioned only in the explanation.
For process components award correct modelling, laws, equations and intermediate
steps actually demonstrated, even when the final answer is wrong. Apply the fixed
error-carried-forward criterion without repeatedly penalizing one arithmetic error.
For final_answer components use the authoritative answer slots, explain conflicts
between slots and explanation. An empty explanation earns no unshown process marks.
Return {"reference_consistent":boolean,"needs_human_review":boolean,"notes":"brief",
"components":[{"name":"exact frozen name","awarded":number,"reason":"specific
evidence in the submitted answer, or what is missing/wrong"}]}.
Include every component exactly once, with 0<=awarded<=maximum. Reference conflicts
or genuinely unresolvable ambiguity require review, not an invented zero or pass.
\end{PhysPrompt}

\end{PhysScopeWideAppendix}


{
  \small
\begin{thebibliography}{27}
\providecommand{\natexlab}[1]{#1}
\providecommand{\url}[1]{\texttt{#1}}
\expandafter\ifx\csname urlstyle\endcsname\relax
  \providecommand{\doi}[1]{doi: #1}\else
  \providecommand{\doi}{doi: \begingroup \urlstyle{rm}\Url}\fi

\bibitem[DeepSeek-AI(2026)]{deepseekai2026deepseekv41flash}
DeepSeek-AI.
\newblock Deepseek-v4.1-flash: Pushing the limits of kv cache compression,
  2026.

\bibitem[Doyle(1979)]{doyle1979truth}
Jon Doyle.
\newblock A truth maintenance system.
\newblock \emph{Artificial intelligence}, 12\penalty0 (3):\penalty0 231--272,
  1979.

\bibitem[{Google DeepMind}(2026)]{googledeepmind2026gemini38flash}
{Google DeepMind}.
\newblock {Gemini 3.8 Flash}: Model card.
\newblock Google DeepMind Model Card, 2026.
\newblock Published September 2, 2026. Accessed September 16, 2026.

\bibitem[Gupta and Kembhavi(2023)]{gupta2023visual}
Tanmay Gupta and Aniruddha Kembhavi.
\newblock Visual programming: Compositional visual reasoning without training.
\newblock In \emph{2023 IEEE/CVF Conference on Computer Vision and Pattern
  Recognition (CVPR)}, pages 14953--14962. IEEE, 2023.

\bibitem[Han et~al.(2026)Han, Hu, Qiu, Wu, and He]{vista2026}
Qiushi Han, Keya Hu, Linlu Qiu, Cathy Wu, and Kaiming He.
\newblock {VISTA}: A visual harness for reasoning in an interactive world,
  2026.

\bibitem[Hu et~al.(2024)Hu, Shi, Fu, Roth, Ostendorf, Zettlemoyer, Smith, and
  Krishna]{hu2024visual}
Yushi Hu, Weijia Shi, Xingyu Fu, Dan Roth, Mari Ostendorf, Luke Zettlemoyer,
  Noah~A Smith, and Ranjay Krishna.
\newblock Visual sketchpad: Sketching as a visual chain of thought for
  multimodal language models.
\newblock \emph{Advances in Neural Information Processing Systems},
  37:\penalty0 139348--139379, 2024.

\bibitem[Jiang et~al.(2026)Jiang, Guo, Wu, Guo, Zhang, Yao, Wu, You, Tang, Cao,
  Wang, Wang, Celi, and Zhang]{jiang2026remaprestoringperceptualcycle}
Hao Jiang, Zhanyu Guo, Chenwei Wu, Yichen Guo, Qizhe Zhang, Junchi Yao, Jixian
  Wu, Jinhao You, Kai Tang, Jiajun Cao, Tinghao Wang, Mengyu Wang, Leo~Anthony
  Celi, and Shanghang Zhang.
\newblock Remap: Restoring the perceptual cycle with reasoning-time latent
  visual memory, 2026.

\bibitem[Liang et~al.(2026)Liang, Liu, Chen, Liu, Chen, Wang, Liu, and
  Lin]{liang2026physalignbenchmarkevidencegroundedrole}
Kecheng Liang, Haoyang Liu, Zexin Chen, Zirong Liu, Weixing Chen, Qiufeng Wang,
  Yang Liu, and Liang Lin.
\newblock Physalign: A benchmark for evidence-grounded role alignment in
  multimodal physics reasoning, 2026.

\bibitem[Lu et~al.(2024)Lu, Bansal, Xia, Liu, Li, Hajishirzi, Cheng, Chang,
  Galley, and Gao]{lu2024mathvista}
Pan Lu, Hritik Bansal, Tony Xia, Jiacheng Liu, Chunyuan Li, Hannaneh
  Hajishirzi, Hao Cheng, Kai-Wei Chang, Michel Galley, and Jianfeng Gao.
\newblock Mathvista: Evaluating mathematical reasoning of foundation models in
  visual contexts.
\newblock In \emph{International Conference on Learning Representations}, pages
  23439--23554, 2024.

\bibitem[Madaan et~al.(2023)Madaan, Tandon, Gupta, Hallinan, Gao, Wiegreffe,
  Alon, Dziri, Prabhumoye, Yang, et~al.]{madaan2023self}
Aman Madaan, Niket Tandon, Prakhar Gupta, Skyler Hallinan, Luyu Gao, Sarah
  Wiegreffe, Uri Alon, Nouha Dziri, Shrimai Prabhumoye, Yiming Yang, et~al.
\newblock Self-refine: Iterative refinement with self-feedback.
\newblock \emph{Advances in neural information processing systems},
  36:\penalty0 46534--46594, 2023.

\bibitem[{OpenAI}(2026{\natexlab{a}})]{openai2026gpt56sol}
{OpenAI}.
\newblock {GPT-5.6 Sol} model.
\newblock OpenAI API Documentation, 2026{\natexlab{a}}.
\newblock Accessed: 2026-09-15.

\bibitem[{OpenAI}(2026{\natexlab{b}})]{openai2026gpt6astra}
{OpenAI}.
\newblock Gpt-6 astra model.
\newblock OpenAI API Documentation, 2026{\natexlab{b}}.
\newblock Accessed: 2026-09-14.

\bibitem[Qiu et~al.(2025)Qiu, Shi, Juan, Zhao, Geng, Liu, Wang, Wu, and
  Wang]{qiu2025physicssupernovaaiagent}
Jiahao Qiu, Jingzhe Shi, Xinzhe Juan, Zelin Zhao, Jiayi Geng, Shilong Liu,
  Hongru Wang, Sanfeng Wu, and Mengdi Wang.
\newblock Physics supernova: Ai agent matches elite gold medalists at ipho
  2025, 2025.

\bibitem[{Qwen Team}(2026)]{qwen3.5}
{Qwen Team}.
\newblock {Qwen3.5}: Towards native multimodal agents, 2026.

\bibitem[Shao et~al.(2024)Shao, Qian, Xiao, Song, Zong, Wang, Liu, and
  Li]{shao2024visual}
Hao Shao, Shengju Qian, Han Xiao, Guanglu Song, Zhuofan Zong, Letian Wang, Yu
  Liu, and Hongsheng Li.
\newblock Visual cot: Advancing multi-modal language models with a
  comprehensive dataset and benchmark for chain-of-thought reasoning.
\newblock \emph{Advances in Neural Information Processing Systems},
  37:\penalty0 8612--8642, 2024.

\bibitem[Shen et~al.(2025)Shen, Wu, Han, Hsieh, Wang, Zhang, Cheng, Hao, Ni,
  Wang, et~al.]{shen2025phyx}
Hui Shen, Taiqiang Wu, Qi Han, Yunta Hsieh, Jizhou Wang, Yuyue Zhang, Yuxin
  Cheng, Zijian Hao, Yuansheng Ni, Xin Wang, et~al.
\newblock Phyx: Does your model have the" wits" for physical reasoning?
\newblock \emph{arXiv preprint arXiv:2505.15929}, 2025.

\bibitem[Shinn et~al.(2023)Shinn, Cassano, Gopinath, Narasimhan, and
  Yao]{shinn2023reflexion}
Noah Shinn, Federico Cassano, Ashwin Gopinath, Karthik Narasimhan, and Shunyu
  Yao.
\newblock Reflexion: Language agents with verbal reinforcement learning.
\newblock \emph{Advances in neural information processing systems},
  36:\penalty0 8634--8652, 2023.

\bibitem[Sun et~al.(2026)Sun, Zhu, Liu, Liang, Sun, Jia, He, Wu, Li, Wei,
  et~al.]{sun2026canvas}
Lingzhuang Sun, Yuxia Zhu, Ruitong Liu, Hao Liang, Zheng Sun, Caijun Jia,
  Honghao He, Yuchen Wu, Siyuan Li, Jingxuan Wei, et~al.
\newblock Canvas-of-thought: Grounding reasoning via mutable structured states.
\newblock \emph{arXiv preprint arXiv:2602.10494}, 2026.

\bibitem[Sur{\'\i}s et~al.(2023)Sur{\'\i}s, Menon, and
  Vondrick]{suris2023vipergpt}
D{\'\i}dac Sur{\'\i}s, Sachit Menon, and Carl Vondrick.
\newblock Vipergpt: Visual inference via python execution for reasoning.
\newblock In \emph{2023 IEEE/CVF International Conference on Computer Vision
  (ICCV)}, pages 11854--11864. IEEE, 2023.

\bibitem[Wang et~al.(2026{\natexlab{a}})Wang, Guan, Qiu, Li, Gai, Yang, Zhou,
  Zhao, Jiang, and Jiang]{wang2026groundingscoreexplicitvisual}
Junxin Wang, Dai Guan, Weijie Qiu, Zhihang Li, Yongbo Gai, Zhengyi Yang, Mengyu
  Zhou, Erchao Zhao, Xiaoxi Jiang, and Guanjun Jiang.
\newblock Grounding the score: Explicit visual premise verification for
  reliable vision-language process reward models, 2026{\natexlab{a}}.

\bibitem[Wang et~al.(2025{\natexlab{a}})Wang, Su, Liu, Li, Xia, Xiao, Zhang,
  Dai, Chen, Meng, et~al.]{wang2025physunibench}
Lintao Wang, Encheng Su, Jiaqi Liu, Pengze Li, Peng Xia, Jiabei Xiao, Wenlong
  Zhang, Xinnan Dai, Xi Chen, Yuan Meng, et~al.
\newblock Physunibench: an undergraduate-level physics reasoning benchmark for
  multimodal models.
\newblock \emph{arXiv e-prints}, pages arXiv--2506, 2025{\natexlab{a}}.

\bibitem[Wang et~al.(2025{\natexlab{b}})Wang, Gao, Gu, Pu, Cui, Wei, Liu, Jing,
  Ye, Shao, et~al.]{wang2025internvl3_5}
Weiyun Wang, Zhangwei Gao, Lixin Gu, Hengjun Pu, Long Cui, Xingguang Wei,
  Zhaoyang Liu, Linglin Jing, Shenglong Ye, Jie Shao, et~al.
\newblock Internvl3.5: Advancing open-source multimodal models in versatility,
  reasoning, and efficiency.
\newblock \emph{arXiv preprint arXiv:2508.18265}, 2025{\natexlab{b}}.

\bibitem[Wang et~al.(2026{\natexlab{b}})Wang, Hu, Wang, and
  Liu]{wang2026physmrv}
Wenyuan Wang, Lianyu Hu, Hao Wang, and Yang Liu.
\newblock Physmrv: Physical memory retrieval and verification for physics
  plausibility reasoning.
\newblock \emph{arXiv preprint arXiv:2607.10190}, 2026{\natexlab{b}}.

\bibitem[Xiang et~al.(2026{\natexlab{a}})Xiang, Li, Zhang, Huang, Liu, Qu, He,
  Chen, Yuan, Han, et~al.]{xiang2026seephys}
Kun Xiang, Heng Li, Terry~Jingchen Zhang, Yinya Huang, Zirong Liu, Peixin Qu,
  Jixi He, Jiaqi Chen, Yu-Jie Yuan, Jianhua Han, et~al.
\newblock Seephys: Does seeing help thinking?--benchmarking vision-based
  physics reasoning.
\newblock \emph{Advances in Neural Information Processing Systems}, 38,
  2026{\natexlab{a}}.

\bibitem[Xiang et~al.(2026{\natexlab{b}})Xiang, Zhang, Liu, Zhou, Tang, Yu, Lu,
  Huang, Li, Zhang, Liu, Zhang, Fang, Guo, Zhen, Tu, Huang, and
  Liang]{xiang2026seephyspro}
Kun Xiang, Terry~Jingchen Zhang, Zirong Liu, Bokai Zhou, Yueling Tang, Junjie
  Yu, Jiacong Lu, Shangrui Huang, Heng Li, Likui Zhang, Kunkun Liu, Changzheng
  Zhang, Yangle Fang, Boqiang Guo, Hui-Ling Zhen, Dandan Tu, Yinya Huang, and
  Xiaodan Liang.
\newblock Seephys pro: Diagnosing modality transfer and blind-training effects
  in multimodal rlvr for physics reasoning.
\newblock \emph{arXiv preprint arXiv:2605.09266}, 2026{\natexlab{b}}.

\bibitem[Zhang et~al.(2025)Zhang, Dong, Wu, Huang, Jia, Fernando, Shou, Zhang,
  and Liu]{zhang2025physreason}
Xinyu Zhang, Yuxuan Dong, Yanrui Wu, Jiaxing Huang, Chengyou Jia, Basura
  Fernando, Mike~Zheng Shou, Lingling Zhang, and Jun Liu.
\newblock Physreason: A comprehensive benchmark towards physics-based
  reasoning.
\newblock In \emph{Proceedings of the 63rd Annual Meeting of the Association
  for Computational Linguistics (Volume 1: Long Papers)}, pages 16593--16615,
  2025.

\bibitem[Zhang et~al.(2026)Zhang, Guo, Pan, Yu, Zhang, Li, and
  Wang]{zhang2026decoupled}
Ye Zhang, Xuehang Guo, Rui Pan, Pengfei Yu, Denghui Zhang, Manling Li, and
  Qingyun Wang.
\newblock Decoupled physical modeling and execution for physics reasoning.
\newblock In \emph{Proceedings of the 2026 Conference on Empirical Methods in
  Natural Language Processing}, pages 32112--32131, 2026.

\end{thebibliography}
}


{
  \small
\begin{thebibliography}{9}
\providecommand{\natexlab}[1]{#1}
\providecommand{\url}[1]{\texttt{#1}}
\expandafter\ifx\csname urlstyle\endcsname\relax
  \providecommand{\doi}[1]{doi: #1}\else
  \providecommand{\doi}{doi: \begingroup \urlstyle{rm}\Url}\fi

\bibitem[DeepSeek-AI(2026)]{deepseekai2026deepseekv41flash}
DeepSeek-AI.
\newblock Deepseek-v4.1-flash: Pushing the limits of kv cache compression,
  2026.

\bibitem[{Google DeepMind}(2026)]{googledeepmind2026gemini38flash}
{Google DeepMind}.
\newblock {Gemini 3.8 Flash}: Model card.
\newblock Google DeepMind Model Card, 2026.
\newblock Published September 2, 2026. Accessed September 16, 2026.

\bibitem[Liang et~al.(2026)Liang, Liu, Chen, Liu, Chen, Wang, Liu, and
  Lin]{liang2026physalignbenchmarkevidencegroundedrole}
Kecheng Liang, Haoyang Liu, Zexin Chen, Zirong Liu, Weixing Chen, Qiufeng Wang,
  Yang Liu, and Liang Lin.
\newblock Physalign: A benchmark for evidence-grounded role alignment in
  multimodal physics reasoning, 2026.

\bibitem[{OpenAI}(2026{\natexlab{a}})]{openai2026gpt56sol}
{OpenAI}.
\newblock {GPT-5.6 Sol} model.
\newblock OpenAI API Documentation, 2026{\natexlab{a}}.
\newblock Accessed: 2026-09-15.

\bibitem[{OpenAI}(2026{\natexlab{b}})]{openai2026gpt6astra}
{OpenAI}.
\newblock Gpt-6 astra model.
\newblock OpenAI API Documentation, 2026{\natexlab{b}}.
\newblock Accessed: 2026-09-14.

\bibitem[{Qwen Team}(2026)]{qwen3.5}
{Qwen Team}.
\newblock {Qwen3.5}: Towards native multimodal agents, 2026.

\bibitem[Sun et~al.(2026)Sun, Zhu, Liu, Liang, Sun, Jia, He, Wu, Li, Wei,
  et~al.]{sun2026canvas}
Lingzhuang Sun, Yuxia Zhu, Ruitong Liu, Hao Liang, Zheng Sun, Caijun Jia,
  Honghao He, Yuchen Wu, Siyuan Li, Jingxuan Wei, et~al.
\newblock Canvas-of-thought: Grounding reasoning via mutable structured states.
\newblock \emph{arXiv preprint arXiv:2602.10494}, 2026.

\bibitem[Wang et~al.(2025{\natexlab{a}})Wang, Su, Liu, Li, Xia, Xiao, Zhang,
  Dai, Chen, Meng, et~al.]{wang2025physunibench}
Lintao Wang, Encheng Su, Jiaqi Liu, Pengze Li, Peng Xia, Jiabei Xiao, Wenlong
  Zhang, Xinnan Dai, Xi Chen, Yuan Meng, et~al.
\newblock Physunibench: an undergraduate-level physics reasoning benchmark for
  multimodal models.
\newblock \emph{arXiv e-prints}, pages arXiv--2506, 2025{\natexlab{a}}.

\bibitem[Wang et~al.(2025{\natexlab{b}})Wang, Gao, Gu, Pu, Cui, Wei, Liu, Jing,
  Ye, Shao, et~al.]{wang2025internvl3_5}
Weiyun Wang, Zhangwei Gao, Lixin Gu, Hengjun Pu, Long Cui, Xingguang Wei,
  Zhaoyang Liu, Linglin Jing, Shenglong Ye, Jie Shao, et~al.
\newblock Internvl3.5: Advancing open-source multimodal models in versatility,
  reasoning, and efficiency.
\newblock \emph{arXiv preprint arXiv:2508.18265}, 2025{\natexlab{b}}.

\end{thebibliography}
}
\end{document}